\documentclass[pdflatex,sn-mathphys-num]{sn-jnl}

\usepackage[T1]{fontenc}
\usepackage[utf8]{inputenc}
\usepackage{graphicx}
\usepackage{multirow}
\usepackage{amsmath,amssymb,amsfonts}
\usepackage{amsthm}
\usepackage{mathrsfs}
\usepackage[title]{appendix}
\usepackage{xcolor}
\usepackage{textcomp}
\usepackage{manyfoot}
\usepackage{booktabs}
\usepackage{listings}
\usepackage{optidef}
\usepackage{subcaption}
\usepackage{times}
\usepackage{enumerate}
\usepackage{latexsym}
\usepackage{pdfpages}
\usepackage{adjustbox}
\usepackage{enumitem}
\usepackage{array, bigdelim, makecell}
\usepackage{url}
\usepackage{hyperref}
\usepackage{setspace}
\usepackage[lined,boxed,commentsnumbered,procnumbered,ruled]{algorithm2e}
\usepackage{titlesec}
\usepackage{tcolorbox}
\tcbuselibrary{skins,breakable}
\usepackage{algpseudocode}
\usepackage{microtype}

\titleformat{\paragraph}[runin]{\normalfont\bfseries}{\theparagraph}{1em}{}
\makeatletter
\def\@xfootnote[#1]{%
  \protected@xdef\@thefnmark{#1}%
  \@footnotemark\@footnotetext}
\makeatother
 \SetKwFunction{FRecurs}{FnRecursive}%

\begin{document}

\title[When Features Become Instances]{When Features Become Instances: Inverted Contrastive Learning for Unsupervised Feature Selection}

\author*[1]{\fnm{Utsab} \sur{Ghosh}}
\author[1]{\fnm{Roshni} \sur{Chakraborty}}

\affil[1]{\orgname{ABV-Indian Institute of Information Technology and Management, Gwalior}, \country{India}}

\abstract{Unsupervised feature selection (UFS) seeks a compact subset of informative features without access to class labels, making feature utility difficult to define. Existing UFS methods therefore rely on indirect structural criteria, such as similarity preservation, locality, sparsity, cluster geometry, or reconstruction quality. In this paper, we instead study UFS through representation consistency and propose Inverted Contrastive Learning for Unsupervised Feature Selection (ICLFS), a feature-wise contrastive framework that reformulates UFS as a representation learning problem over features rather than samples. ICLFS first inverts the data matrix so that each feature is represented by its sample-profile vector, then constructs multiple masked positive views together with a shuffled negative view, and learns projector-space representations that remain consistent across these structured perturbations under an InfoNCE-based objective. Motivated by recent findings that cosine-based and InfoNCE-based training affect embedding norms, we use projector-space embedding magnitude as the saliency signal for ranking features. To further improve separation among features, we incorporate a decorrelation regularizer that encourages distinct features to occupy more separated regions of the learned embedding space. The resulting norm-based ranking is subsequently refined through Laplacian-Gated Ranking Correction, which suppresses locally redundant candidates while preserving salient ones. Extensive experiments on 12 benchmark datasets show that ICLFS achieves the best clustering accuracy on 10 datasets against both classical and neural baselines under the standard clustering-based UFS evaluation protocol, while remaining competitive on the other two. These results show that feature-wise contrastive representation consistency provides a strong and effective alternative to neighborhood, cluster, and reconstruction-based UFS formulations.}


\maketitle

\section{Introduction}
\par High-dimensional datasets commonly include noisy, redundant, and only weakly
informative features, which can impair clustering performance and complicate
exploratory analysis~\cite{guyon2003introduction,li2017feature}. This issue is
particularly prominent in domains such as biomedicine~\cite{saeys2007review},
text mining~\cite{cai2018feature}, and image retrieval~\cite{cai2018feature},
where the feature dimension can be very large relative to the number of
available samples, and only a fraction of the measured variables may capture
the true structure of the data. In supervised feature selection, labels guide
the identification of predictive features and help eliminate irrelevant or
redundant variables~\cite{guyon2003introduction,saeys2007review}. In many
real-world scenarios, however, labeled data are scarce, while unlabeled data
are much more abundant~\cite{guyon2003introduction}, making direct supervised
selection impractical. Unsupervised feature selection (UFS) addresses this
setting by selecting a compact subset of informative features without relying
on label information, thereby improving interpretability, reducing redundancy,
and facilitating downstream learning~\cite{li2017feature}.

\par As no label signal is available, UFS is commonly formulated through
indirect structural criteria rather than task-level supervision. Existing
methods typically assess feature usefulness through properties such as
similarity preservation, locality, sparsity, or cluster
geometry~\cite{cai2018feature}. As a result, feature relevance must be
inferred from the intrinsic organization of the data itself, making robust
feature identification considerably more challenging in high-dimensional
settings~\cite{li2017feature,cai2018feature,garcia2016high,urbanowicz2018relief}. Existing research works in this
area can be broadly grouped according to the structural principles used in
their proposed approaches. For example, filter-style methods such as
Laplacian Score (LS) rank features according to their ability to preserve
local neighborhood relations~\cite{he2005laplacian}. Spectral Feature
Selection (SPEC) evaluates features within a spectral graph-theoretic
framework derived from structural similarity~\cite{zhao2007spec}, whereas
Multi-Cluster Feature Selection (MCFS) emphasizes preservation of latent
multi-cluster structure~\cite{cai2010unsupervised}. Subsequently, several
research works focus on recovering more discriminative unlabeled structure.
For instance, Nonnegative
Discriminative Feature Selection (NDFS) jointly learns latent cluster
assignments and sparse feature weights through nonnegative spectral
analysis~\cite{li2012unsupervised}. More recently, neural approaches such as
Concrete Autoencoders (CAE) have framed feature selection through
reconstruction objectives, selecting subsets of features that best reconstruct
the original input~\cite{balin2019concrete} and LS-CAE ~\cite{shaham2022deep}. Despite their empirical effectiveness, these approaches mainly focus on surrogate principles, such as, neighborhood preservation, latent cluster
recovery, or reconstruction quality. In contrast we assess feature utility
through the consistency of learned feature representations across masked
feature views, i.e., systematically constructed views that partially mask the
observed feature profile while preserving the identity of the underlying
feature. 


\par This motivates studying UFS through representation consistency rather
than through a single predefined structural heuristic. In this paper, we
approach UFS from a feature-wise contrastive learning perspective, where each
feature is paired with multiple masked positive views and optimized through an
InfoNCE-based objective in the learned representation
space~\cite{oord2018representation}. However, Wang et
al.~\cite{wang2017normface} and Zhang et al.~\cite{zhang2020spherical} showed
that cosine-based training affects embedding norms and that these norms in
turn affect gradient magnitudes. More recently, Draganov et
al.~\cite{draganovimportance} showed that optimizing InfoNCE increases these
embedding norms across self-supervised learning methods and argued that this
may occur when the network treats frequently reinforced samples as more
prototypical. Additionally, embedding magnitude has been observed to serve as
a measure of model certainty~\cite{scott2021von,kirchhof2023probabilistic}.
Therefore, in this paper, we propose projector-space embedding magnitude
as a saliency signal for unsupervised feature selection. Based on this idea, we
propose an Inverted Contrastive Learning for Unsupervised Feature Selection  (ICLFS), a
feature-wise contrastive framework that reformulates UFS as a representation
learning problem over features rather than samples. ICLFS first inverts the
data matrix so that each feature is represented by its sample-profile vector,
then constructs multiple masked positive views and learns projector-space
representations that remain consistent across them under an InfoNCE-based \cite{oord2018representation}
objective. To further improve separation among features, we also include
a decorrelation regularizer that encourages distinct features to occupy more
separated regions of the learned embedding space. After optimization, features are ranked according to the norm of their projector-space embeddings, and this
ranking is subsequently refined through Laplacian-Gated Ranking Correction
that suppresses locally redundant candidates while preserving highly salient
ones.

\par Our exhaustive empirical evaluation shows that ICLFS achieves the best
clustering accuracy on 10 of the 12 benchmark datasets against both classical
and neural baselines under the standard clustering-based UFS evaluation
protocol~\cite{shaham2022deep}. Our empirical evaluations indicate that ICLFS
outperforms existing methods in k-means clustering accuracy by
0.47\% to 2.74\% on ARCENE, 0.44\% to 1.20\% on BASEHOCK, 1.85\% to 12.37\%
on COIL20, 0.44\% to 19.91\% on GISETTE, 8.40\% to 14.24\% on LUNG,
11.25\% to 15.16\% on NCI9, 0.94\% to 1.64\% on PCMAC, 23.19\% to 27.89\%
on PROSTATE, 4.57\% to 7.69\% on TOX171, and 2.81\% to 15.78\% on
WARPPIE10P. We observe that ICLFS achieves the largest gains on PROSTATE,
with an improvement range of 23.19\% to 27.89\%, followed by GISETTE,
WARPPIE10P, NCI9, and LUNG. On the two datasets where ICLFS does not achieve
the top result, the margin relative to the baseline range is -4.51\% to
12.09\% on ALLAML and -0.41\% to 1.23\% on RELATHE, indicating that RELATHE
remains highly competitive whereas LS-CAE is the stronger feature selector
for ALLAML.

The main contributions of this paper are as follows:
\begin{itemize}
\item We introduce Inverted Contrastive Learning with Feature Screening
(ICLFS), a feature-wise unsupervised feature selection framework that
reformulates UFS as a contrastive representation learning problem by treating
features, rather than samples, as the learning instances.

\item We develop a ranking mechanism based on projector-space embedding norms,
together with a post hoc Laplacian-Gated Ranking Correction stage, to produce
a global feature ranking that can be truncated to obtain feature subsets at
different target cardinalities.

\item We show that the proposed framework performs competitively against both
classical and neural UFS baselines across a diverse benchmark suite, achieving
the best clustering accuracy on 10 of 12 datasets under the standard
clustering-based evaluation protocol. On the remaining two datasets, ICLFS
achieves a tied second-best result on RELATHE and remains competitive on
ALLAML, where the strongest performance is obtained by LS-CAE.

\item We provide empirical evidence that feature-wise contrastive
representation consistency offers an effective alternative to neighborhood-,
cluster-, and reconstruction-based optimization proxies for unsupervised
feature selection.
\end{itemize}

\par The remainder of this paper is organized as follows. Section \ref{sec:related_work} reviews the
most relevant prior work in unsupervised feature selection and related
representation-learning directions. Section \ref{sec:core} presents the proposed ICLFS
framework, including its feature-wise formulation, model architecture,
optimization objective, and redundancy-aware refinement stage. Section \ref{sec:experiments}
describes the benchmark datasets, evaluation protocol, and implementation
details. Section \ref{sec:results} reports the comparative, ablation, and analytical results.
Finally, Section \ref{sec:conclusion} concludes the paper and outlines directions for future work.\par The implementation associated with this work is publicly available at:
\url{https://github.com/neilghos/ICLFS}.

\section{Related Work}
\label{sec:related_work}
Unsupervised feature selection (UFS) aims to identify a compact subset of
informative features without access to class labels. Unlike supervised feature
selection, where discriminative relevance can be measured directly against a
target variable, UFS must rely on indirect structural principles such as local
manifold preservation, cluster consistency, sparsity, or reconstruction quality.
As a result, the design of effective UFS methods is closely tied to the choice
of surrogate objective used to approximate feature utility. Existing work in
this area can be broadly grouped into classical graph- and spectral-based
methods, reconstruction-driven deep models, and more recent representation
learning approaches.

\subsection{Classical Unsupervised Feature Selection}

A large body of early UFS work is based on graph structure and spectral
analysis. Laplacian Score (LS) ranks features according to how well they
preserve local neighborhood structure, under the intuition that informative
features should respect the intrinsic manifold geometry of the data
\cite{he2005laplacian}. Spectral Feature Selection (SPEC) extends this
perspective by evaluating features through a spectral graph-theoretic framework
that measures consistency with the global structure encoded by the similarity
graph \cite{zhao2007spec}. Multi-Cluster Feature Selection (MCFS) further
develops this line of work by selecting features that best preserve latent
multi-cluster structure through spectral embedding and sparse regression
\cite{cai2010unsupervised}. These methods are influential and often effective,
but they are fundamentally tied to predefined similarity graphs or spectral
surrogates, which may not fully capture richer feature interactions.

Another important classical direction seeks to learn feature importance jointly
with latent discriminative or clustering structure. Unsupervised
Discriminative Feature Selection (UDFS) follows this idea by combining
discriminative analysis with sparsity regularization in a label-free setting,
thereby encouraging the selection of features that preserve latent
separability without access to class labels. Nonnegative Discriminative
Feature Selection (NDFS) further combines spectral clustering with
nonnegative sparse feature selection, allowing cluster structure and feature
selection to reinforce one another~\cite{li2012unsupervised}. Such methods
are generally more adaptive than simple filter-based ranking schemes, but
they still remain closely coupled to cluster-recovery objectives and may
become sensitive to graph quality or optimization difficulty in
high-dimensional settings.

\subsection{Deep and Reconstruction-Based Feature Selection}

With the rise of deep learning, reconstruction-based and differentiable
frameworks became a major direction for UFS. Concrete Autoencoders (CAE) learn
a discrete feature subset through differentiable subset selection, optimizing
the selected features for reconstruction of the original
input~\cite{balin2019concrete}. Around the same period, Differentiable
Unsupervised Feature Selection (DUFS) \cite{lindenbaum2021differentiable} further reflected the move toward
differentiable UFS by introducing a gated Laplacian-based objective that
learns feature selection within a continuous optimization framework. More
recently, LS-CAE enriched the reconstruction-based paradigm by combining
autoencoder reconstruction with additional mechanisms designed to discard
nuisance and correlated features while preserving structural
information~\cite{shaham2022deep}.

These methods are often strong empirical baselines, but their primary learning
signal remains tied to reconstruction, surrogate prediction, or differentiable
selection objectives rather than to feature-level relational consistency. As a
result, they may preserve features that are useful for input recovery or proxy
optimization without necessarily emphasizing those that are most informative
for downstream partition structure or non-redundant relational organization.

More broadly, deep UFS methods based on autoencoders, sparse bottlenecks, or
embedded selectors typically focus on compressive representation quality rather
than explicitly modeling feature-to-feature contrastive relations. This leaves
a gap between reconstruction-oriented selection and relational feature
ranking, especially in settings where the key challenge is not input recovery
but the identification of discriminative, complementary, and cluster-relevant
features. A recent related direction is Spectral Self-supervised Feature
Selection (SSFS)~\cite{segalspectral}, which combines graph spectral structure
with self-supervised surrogate learning for feature ranking. While distinct in
formulation, it likewise illustrates the broader move beyond fixed classical
filters toward more adaptive unsupervised feature selection objectives.
\subsection{Contrastive and Self-Supervised Representation Learning}

Contrastive learning has become a central paradigm in self-supervised
representation learning, particularly in domains where multiple views of the
same instance can be constructed and used to learn invariant embeddings.
However, most contrastive methods are formulated at the \emph{sample level}:
individual data points are treated as instances, augmented sample views are
contrasted, and the resulting representation space is optimized for instance
discrimination. This standard setup does not directly address the central goal
of unsupervised feature selection, where the objective is to assess the utility
of \emph{features} rather than samples.

Only limited attention has been given to adapting contrastive principles to a
feature-centric unsupervised feature selection setting. In particular, existing
approaches rarely invert the data matrix and treat features as learning
instances in their own right. As a result, most contrastive objectives
remain misaligned with the core requirement of UFS: assigning meaningful
saliency to features based on their stability and informativeness under
multiple unsupervised views. This creates an opening for a feature-wise
contrastive formulation in which representation consistency is used as the
basis for feature ranking.

Recent work further strengthens this motivation by showing that contrastive
optimization can induce informative behavior in embedding norms even when
similarity is computed on normalized representations. Wang et
al.~\cite{wang2017normface} and Zhang et al.~\cite{zhang2020spherical} showed
that cosine-based training affects embedding norms and that these norms
influence gradient magnitudes during optimization. More recently, Draganov et
al.~\cite{draganovimportance} showed that optimizing InfoNCE systematically
increases embedding norms across self-supervised learning methods and argued
that this behavior may arise when frequently reinforced samples are treated as
more prototypical in the learned representation space. Related work has also
connected embedding magnitude to confidence- and certainty-related behavior in
representation learning~\cite{scott2021von,kirchhof2023probabilistic}. Taken
together, these observations suggest that projector-space embedding norm can
serve not only as a training byproduct, but also as a potentially meaningful
signal for feature saliency in a feature-wise contrastive learning framework.

\subsection{Feature Interaction Modeling and Redundancy Control}

A persistent challenge in UFS is that informative features must be not only
individually useful but also collectively non-redundant. In classical methods,
redundancy is often handled indirectly through graph-based structure
preservation, spectral objectives, or sparsity-inducing formulations, which aim
to discourage the selection of features that contribute overlapping structural
information. In more recent neural approaches, redundancy control is typically
introduced through the learning objective itself. CAE relies primarily on
reconstruction quality, LS-CAE supplements reconstruction with additional
structure-aware criteria to suppress nuisance and correlated features, and DUFS
uses a differentiable gated Laplacian objective to regularize selection within
a continuous optimization framework.

Despite these advances, explicit redundancy control is still usually embedded
inside a particular scoring or training objective rather than paired with a
separate deterministic refinement stage after representation learning. As a
result, existing methods seldom combine learned feature representations,
structured multi-view perturbation, explicit inter-feature interaction
modeling, and post hoc redundancy-aware filtering within a single coherent
framework. This is particularly limiting in high-dimensional settings, where
strongly ranked features may still be highly correlated, and where a learned
representation space may benefit from an additional refinement step that
suppresses locally redundant selections without discarding globally salient
features.

\subsection{Positioning of the Proposed Method}

The proposed ICLFS framework is motivated by the above limitations. Unlike
classical graph-based ranking methods, ICLFS does not rely solely on fixed
similarity proxies or spectral consistency criteria. Unlike
reconstruction-driven autoencoder approaches, it is not trained primarily to
recover the input. Instead, it formulates unsupervised feature selection as a
feature-level contrastive representation learning problem by inverting the data
matrix and treating each feature profile as an individual learning instance.
Structured multi-view masking supplies complementary positive views, the
interaction module models dependencies across features, and a Laplacian-based
refinement stage suppresses locally redundant selections in the final ranked
set. In this way, ICLFS unifies contrastive representation learning,
inter-feature interaction modeling, and redundancy-aware refinement within a
single UFS framework.

\section{Proposed Approach}
\label{sec:core}
In this section, we present the proposed ICLFS framework for unsupervised
feature selection. We first formalize the problem setting and summarize the
overall pipeline, and then describe the four main components of the framework, i.e., preprocessing and feature-wise inversion, contrastive view construction,
the neural model together with its optimization strategy, followed by the final
selection and refinement stage.

\subsection{Problem Statement}
\label{subsec:problem_statement}
This subsection formalizes the unsupervised feature selection problem that the
proposed framework aims to solve. It also clarifies the selection objective
that motivates the subsequent feature-wise contrastive formulation.

Let the unlabeled data matrix be denoted by
\begin{equation}
\mathbf{X} \in \mathbb{R}^{n \times d},
\label{eq:problem_data}
\end{equation}
where \(n\) is the number of samples and \(d\) is the number of features.
Rows of \(\mathbf{X}\) correspond to samples and columns correspond to measured
variables. In unsupervised feature selection, the objective is to identify a
subset of feature indices
\begin{equation}
\mathcal{S}_k \subseteq \{1, \dots, d\}, \qquad |\mathcal{S}_k| = k,
\label{eq:problem_subset}
\end{equation}
such that the resulting reduced data matrix
\begin{equation}
\mathbf{X}_{\mathcal{S}_k} \in \mathbb{R}^{n \times k},
\label{eq:problem_reduced}
\end{equation}
retains the most informative structure of the original data according to an
unsupervised selection criterion. As shown in
Eq.~\ref{eq:problem_subset}, the task is to select exactly \(k\) features
from the full feature index set, and Eq.~\ref{eq:problem_reduced} gives the
corresponding reduced representation. Since no class labels are available, this structure must be inferred from the
organization of the data itself, such as feature dependencies, sample
similarity, local neighborhood structure, or learned representation
consistency. Rather than learning a separate model for each target
cardinality, we learn a single feature-ordering mechanism from which the
subset \(\mathcal{S}_k\) in Eq.~\ref{eq:problem_subset} can be obtained for
any target cardinality \(k\). In the proposed framework, this is achieved by
first learning an ordering over features and then refining
cardinality-specific selections from that ordering. We next summarize the
proposed framework used to obtain these subsets.

\begin{figure*}[t]
    \centering
    \includegraphics[width=\textwidth]{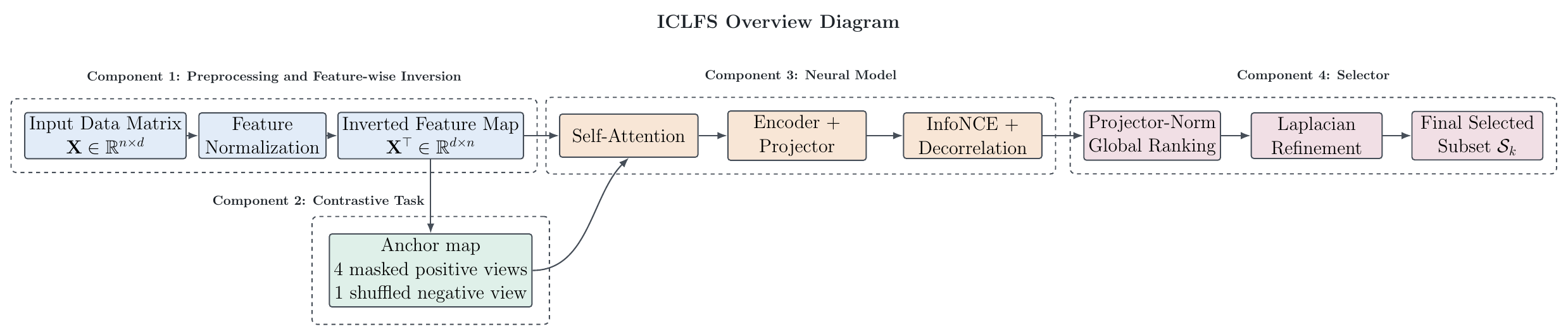}
    \caption{Component-wise overview of the proposed Inverted Contrastive
    Learning for Unsupervised Feature Selection (ICLFS) framework.
    [Component~1: Preprocessing and Feature-wise Inversion] normalizes the input matrix
    and inverts it into a feature-wise map. [Component~2: Contrastive Task]
    generates the masked positive views and shuffled negative view that define
    the unsupervised contrastive objective. [Component~3: Neural Model]
    processes the anchor map and generated views through the self-attention,
    encoder, and projector modules and optimizes the resulting embeddings using
    the InfoNCE objective together with the decorrelation regularizer.
    [Component~4: Selector] ranks features by projector-space norm and refines
    the top-ranked candidates through Laplacian-Gated Ranking Correction
    (LGRC) to produce the final selected subset.}
    \label{fig:ICLFS-overview}
\end{figure*}

\subsection{Overview}
\label{subsec:overview}

This subsection provides a high-level overview of the proposed framework used
to learn the feature-scoring and selection mechanism introduced in the problem
statement. As shown in Fig.~\ref{fig:ICLFS-overview}, the framework is
organized into four components: Component I (Preprocessing and Feature-wise
Inversion), Component II (Contrastive Task), Component III (Neural Model), and
Component IV (Selector). Component I transforms the input data matrix
\(\mathbf{X}\) into the inverted feature map \(\mathbf{X}^{\top}\), in which
each feature is represented across the full sample set. Component II constructs
multiple masked positive feature maps together with a shuffled negative feature
map and stacks them into the generated feature-map tensor \(\mathbf{M}\) to
define the unsupervised contrastive task. Component III processes the anchor
feature map \(\mathbf{X}^{\top}\) and the generated feature-map tensor
\(\mathbf{M}\) through self-attention, nonlinear encoding, projection, and
optimization to learn feature-wise contrastive embeddings. Finally, Component
IV derives an initial projector-norm ranking from the stacked projector output
\(\mathcal{Z}\) and applies Laplacian-Gated Ranking Correction (LGRC) as a
last-mile refinement to obtain the final selected subset. In the next
subsection, Section~\ref{subsec:Component1}, we begin with Component I
(Preprocessing and Feature-wise Inversion).

\subsection{Component I: Preprocessing and Feature-wise Inversion}
\label{subsec:Component1}
This component establishes the input representation on which the rest of the
proposed framework is built. Since ICLFS defines the contrastive task over
features rather than over samples, the sample-wise data matrix
\(\mathbf{X} \in \mathbb{R}^{n \times d}\) introduced in
Eq.~\ref{eq:problem_data} must first be preprocessed and then converted into a
feature-wise representation in which each feature becomes a learning instance.

As defined in Eq.~\ref{eq:problem_data}, \(\mathbf{X}\) is represented in the
usual sample-wise form, with rows corresponding to samples and columns
corresponding to features. To reduce scale imbalance across the feature
columns of \(\mathbf{X}\) and to prevent subsequent representation learning
from being dominated by variables with larger raw magnitudes, we standardize
\(\mathbf{X}\) across samples so that each feature has zero mean and unit
variance. After preprocessing, we transpose the input data matrix \(\mathbf{X}\) in
Eq.~\ref{eq:problem_data} to obtain the inverted feature map
\begin{equation}
\mathbf{X}^{\top} \in \mathbb{R}^{d \times n},
\label{eq:inverted_map}
\end{equation}
as shown in Eq.~\ref{eq:inverted_map}. In the inverted feature map
\(\mathbf{X}^{\top}\), each row corresponds to one feature observed across all
\(n\) samples.Accordingly, the \(j\)-th row of \(\mathbf{X}^{\top}\) is written as
\begin{equation}
\mathbf{x}_j \in \mathbb{R}^{n},
\label{eq:feature_profile}
\end{equation}
where Eq.~\ref{eq:feature_profile} defines the sample-profile vector of
feature \(j\), which serves as one feature instance in the proposed
framework.

\begin{figure*}[t]
    \centering
    \includegraphics[width=\textwidth]{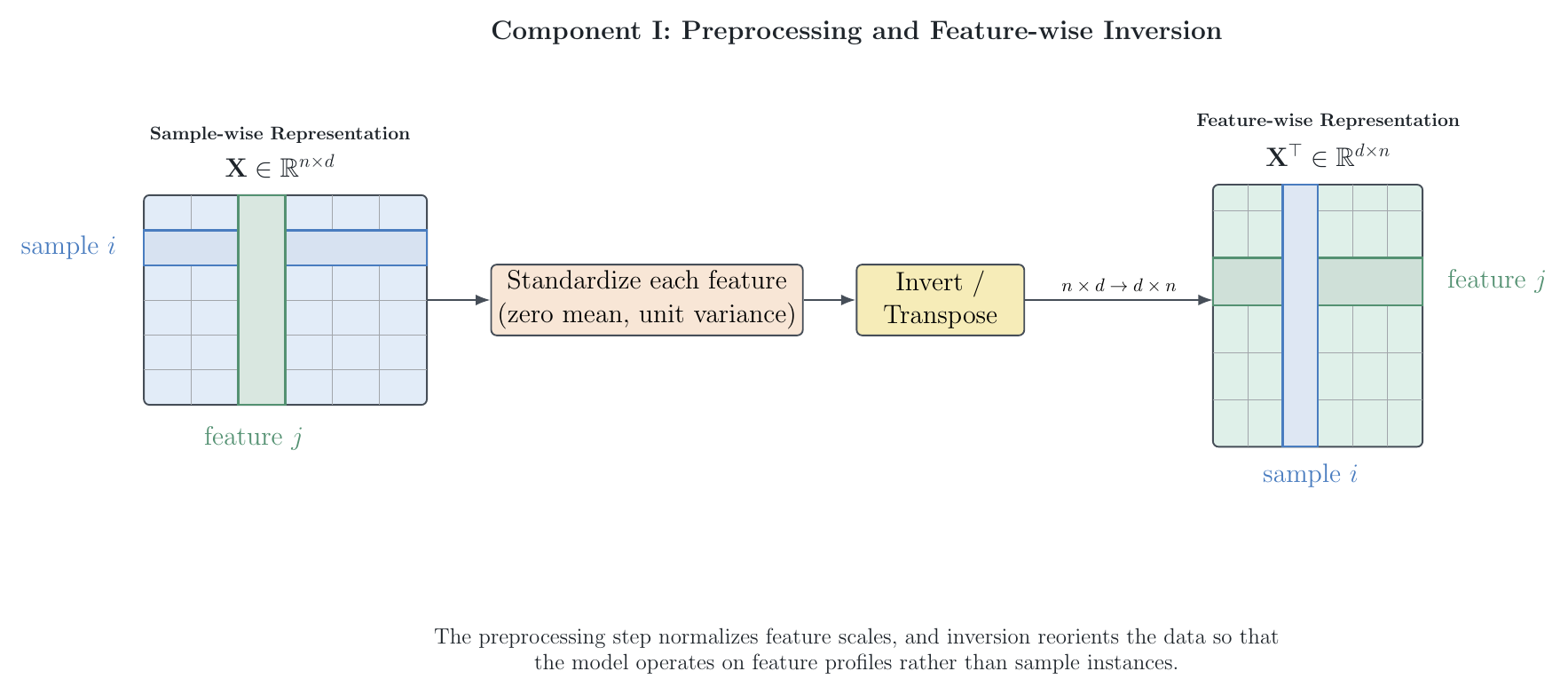}
    \caption{Component I: preprocessing and feature-wise inversion. The sample-wise
data matrix \(\mathbf{X} \in \mathbb{R}^{n \times d}\) is standardized and
transposed to obtain the inverted feature map
\(\mathbf{X}^{\top} \in \mathbb{R}^{d \times n}\), where each row represents
one feature across the full sample set.}
    \label{fig:Component1}
\end{figure*}

As illustrated in Fig.~\ref{fig:Component1}, the preprocessing and inversion
step converts the sample-wise data matrix \(\mathbf{X}\) in
Eq.~\ref{eq:problem_data} into the inverted feature map
\(\mathbf{X}^{\top}\) in Eq.~\ref{eq:inverted_map}. This inverted feature map
\(\mathbf{X}^{\top}\) serves as the input representation of the proposed
framework, allowing all features to be processed jointly within a single batch
so that the subsequent interaction module can capture dependencies among them
before contrastive optimization is applied. Based on the feature-wise
representation defined by Eq.~\ref{eq:inverted_map} and
Eq.~\ref{eq:feature_profile}, the proposed framework is designed to learn a
feature-scoring and ranking mechanism from which the selected subset
\(\mathcal{S}_k\) for any target cardinality \(k\) is obtained, consistent
with the problem formulation in Eq.~\ref{eq:problem_subset}. We next construct
the masked and shuffled views that instantiate the contrastive task.

\begin{figure*}[t]
    \centering
    \includegraphics[width=\textwidth]{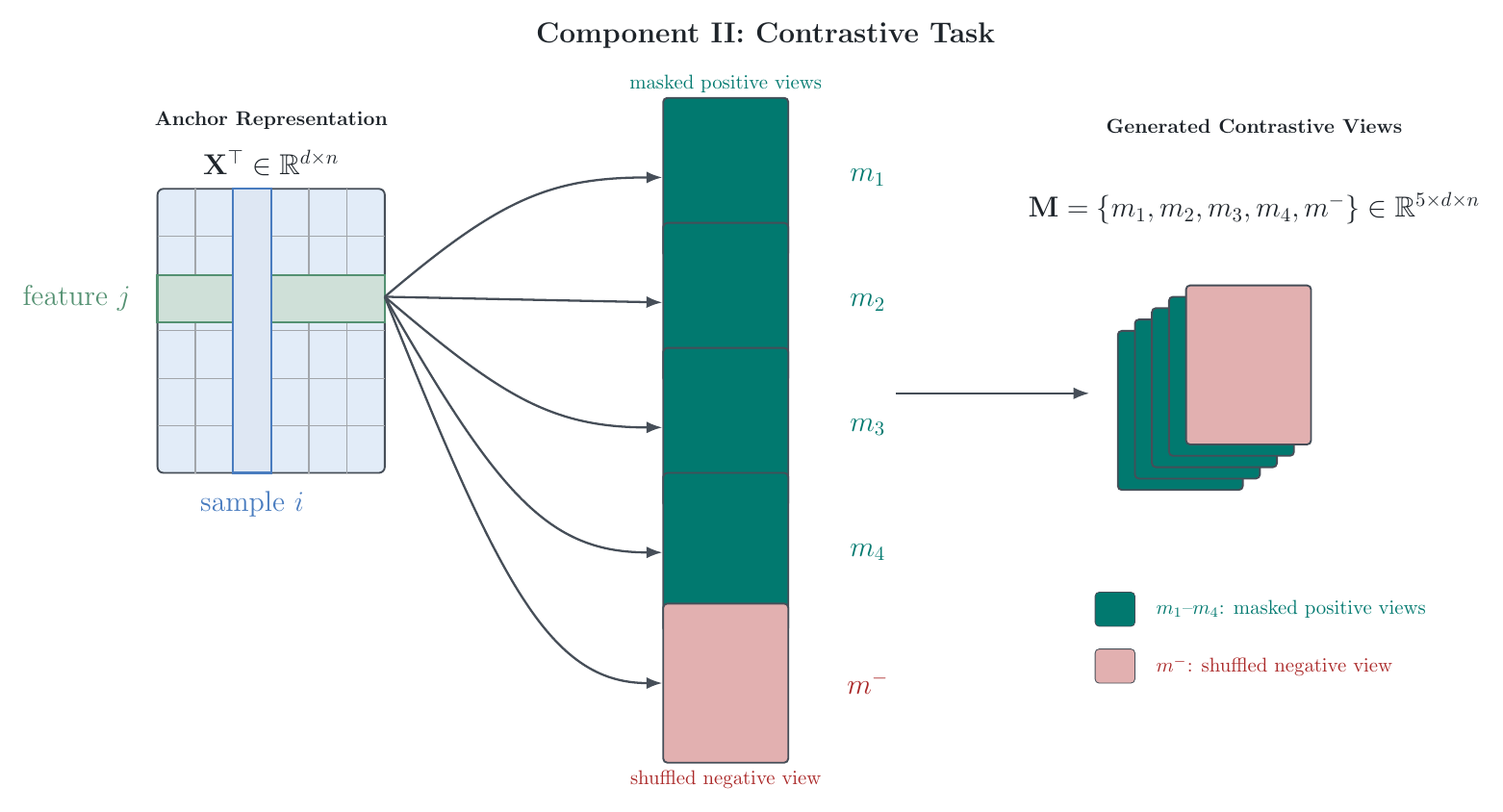}
    \caption{Component II: contrastive task. Starting from the anchor feature map
\(\mathbf{X}^{\top}\), four masked positive feature maps and one shuffled
negative feature map are generated and stacked to define the feature-wise
contrastive view set used for training.}
    \label{fig:Component2}
\end{figure*}

\subsection{Component II: Contrastive Task}
\label{subsec:pert}
This component defines the unsupervised contrastive task used to learn feature
saliency from the feature-wise representation introduced in
Component~I. Figure~\ref{fig:Component2} summarizes the corresponding
view-generation process. The task is designed to create multiple positive
observations of the same feature together with a mismatched observation, so
that the model can identify features whose representations remain stable under
structured perturbation without using class labels or predefined targets. This
design is also motivated by recent findings on InfoNCE-based representation
learning. In particular, Draganov et al.~\cite{draganovimportance} showed that
InfoNCE optimization can increase embedding norms, particularly for instances
that are more frequently reinforced or that occupy denser regions of the
learned representation space.

In the proposed framework, the positive views are constructed as masked
versions of the same feature so that they preserve its identity under partial
observation, whereas the negative view is constructed by shuffling the sample
order of that same feature so that it disrupts the original sample-aligned
structure while remaining derived from the same underlying signal. This makes
the explicit negative more comparable to the anchor than an unrelated feature
would be, because the contrast is formed between structure-preserving and
structure-disrupting transformations of the same feature rather than between
two different features that may differ for many unrelated reasons. Features
that remain stable across the masked positive views therefore receive more
consistent contrastive reinforcement, which motivates the later use of
projector-space embedding norm as a feature saliency signal.

To construct the positive and negative views, we take the inverted feature map
\(\mathbf{X}^{\top}\) defined in Eq.~\ref{eq:inverted_map} as the anchor
feature map. From this anchor feature map, four masked feature maps
\(\mathbf{m}_1,\mathbf{m}_2,\mathbf{m}_3,\mathbf{m}_4\) are constructed as
positive views, while the shuffled feature map \(\mathbf{m}^{-}\) is
constructed as an explicit negative view. The remaining feature instances
within each feature map act as implicit negatives through the contrastive
objective. This yields the feature-level contrastive formulation used in the
proposed method.

Formally, starting from the anchor feature map \(\mathbf{X}^{\top}\) defined
in Eq.~\ref{eq:inverted_map}, we generate four masked positive feature maps
and one shuffled negative feature map,
\begin{equation}
\mathbf{m}_1,\mathbf{m}_2,\mathbf{m}_3,\mathbf{m}_4,\mathbf{m}^{-}
\in \mathbb{R}^{d \times n}.
\label{eq:generated_maps}
\end{equation}
As shown in Eq.~\ref{eq:generated_maps}, all generated feature maps retain the
same \(d \times n\) structure as the anchor feature map in
Eq.~\ref{eq:inverted_map}. The generated feature maps are defined as follows:
\begin{itemize}
\item \(\mathbf{m}_1\): a light-mask feature map obtained by applying a light masking operation to each feature instance in \(\mathbf{X}^{\top}\),
\item \(\mathbf{m}_2\): a heavy-mask feature map obtained by applying a heavier masking operation to each feature instance in \(\mathbf{X}^{\top}\),
\item \(\mathbf{m}_3\): a complementary masked feature map constructed from one masked subset of each feature instance in \(\mathbf{X}^{\top}\),
\item \(\mathbf{m}_4\): a second complementary masked feature map constructed with controlled overlap relative to \(\mathbf{m}_3\),
\item \(\mathbf{m}^{-}\): a shuffled negative feature map obtained by randomly permuting the sample order within each feature instance in \(\mathbf{X}^{\top}\).
\end{itemize}

The masked feature maps preserve the identity of each underlying feature under
partial observation, whereas the shuffled feature map disrupts the original
sample-aligned structure of every feature instance while preserving its marginal
values, thereby providing an explicit mismatched view. The exact masking
ratios and overlap settings used to construct these feature maps are given in
Section~\ref{sec:implementation_details}. The generated feature maps are stacked as
\begin{equation}
\mathbf{M}
=
\left[
\mathbf{m}_1,\mathbf{m}_2,\mathbf{m}_3,\mathbf{m}_4,\mathbf{m}^{-}
\right]
\in \mathbb{R}^{5 \times d \times n},
\label{eq:stacked_views}
\end{equation}
as illustrated in Fig.~\ref{fig:Component2}. Thus, Eq.~\ref{eq:stacked_views}
collects all generated feature maps into a single stacked generated
feature-map tensor that is used in the subsequent neural model. For the
\(j\)-th feature, the corresponding row-wise views are
\(\mathbf{m}_1[j,:]\), \(\mathbf{m}_2[j,:]\), \(\mathbf{m}_3[j,:]\),
\(\mathbf{m}_4[j,:]\), and \(\mathbf{m}^{-}[j,:]\), which define the
positive and negative observations associated with the anchor feature vector
\(\mathbf{x}_j\) in Eq.~\ref{eq:feature_profile}. We next describe how the
anchor feature map \(\mathbf{X}^{\top}\) and the stacked generated
feature-map tensor \(\mathbf{M}\) are processed by the neural model.

\subsection{Component III: Neural Model}
This component describes the neural model that transforms the anchor feature
map \(\mathbf{X}^{\top}\) in Eq.~\ref{eq:inverted_map} and the generated
feature-map tensor \(\mathbf{M}\) in Eq.~\ref{eq:stacked_views} into
feature-wise encoder and projector embeddings. Specifically, the neural model
consists of an encoder that produces latent feature embeddings, a projector
that maps these embeddings into the contrastive space, and the model-scaling
and optimization choices used to train them.

\subsubsection{Encoder}
\label{subsubsec:encoder}
This subsection describes the encoder used to transform the anchor feature map
\(\mathbf{X}^{\top}\) and the generated feature maps in \(\mathbf{M}\) into
latent feature embeddings. In high-dimensional data, the usefulness of a
feature is often not purely an individual property, but may also depend on
its relation to the remaining features in the dataset. For this reason, in
ICLFS the encoder first applies a self-attention layer to model inter-feature
relations. The resulting feature map is then transformed by nonlinear layers
into an encoder embedding matrix.

Let
\begin{equation}
\mathbf{V} \in \mathbb{R}^{d \times n},
\label{eq:encoder_input_map}
\end{equation}
denote an input feature map, where \(\mathbf{V}\) may be the anchor feature
map \(\mathbf{X}^{\top}\) or one of the generated feature maps contained in
\(\mathbf{M}\). The encoder first applies a self-attention layer across the
rows of \(\mathbf{V}\); in our implementation, this layer uses a single
attention head. In this formulation, each row corresponds to one feature
instance, and its \(n\)-dimensional sample-profile vector serves as the
corresponding representation. This allows each feature instance to attend to
all other feature instances within the same feature map before row-wise
encoding is applied.

The attention-mixed feature map is then passed through a two-stage multilayer
perceptron encoder, applied row-wise. The first encoder linear layer maps each
feature instance into an intermediate hidden representation of dimension \(d_e\),
followed by batch normalization, LeakyReLU activation, and dropout. This
intermediate hidden matrix is written as
\begin{equation}
\mathbf{E}^{(\mathbf{V})} \in \mathbb{R}^{d \times d_e},
\label{eq:encoder_hidden_matrix}
\end{equation}
The second encoder linear layer then maps \(\mathbf{E}^{(\mathbf{V})}\) into
the encoder embedding matrix shown in
Eq.~\ref{eq:encoder_embedding_matrix}:
\begin{equation}
\mathbf{H}^{(\mathbf{V})} \in \mathbb{R}^{d \times d_h},
\label{eq:encoder_embedding_matrix}
\end{equation}
where \(d_h\) denotes the encoder embedding dimension, and the \(j\)-th row
\begin{equation}
\mathbf{h}^{(\mathbf{V})}_j \in \mathbb{R}^{d_h}
\label{eq:encoder_row_embedding}
\end{equation}
denotes the final encoder embedding of feature \(j\) for the input feature map
\(\mathbf{V}\). Thus, each input feature map is first mapped to the
intermediate hidden matrix \(\mathbf{E}^{(\mathbf{V})}\) and then to the final
embedding matrix \(\mathbf{H}^{(\mathbf{V})}\), whose rows represent the
latent embeddings of all \(d\) features.

When the encoder is applied to the anchor feature map \(\mathbf{X}^{\top}\),
the resulting encoder embedding matrix is
\begin{equation}
\mathbf{H}^{(a)} \in \mathbb{R}^{d \times d_h}.
\label{eq:anchor_embedding_matrix}
\end{equation}
Likewise, when the encoder is applied to each generated feature map in
\(\mathbf{M}\), it produces the generated encoder embedding collection
\begin{equation}
\bar{\mathbf{H}}
=
\left[
\mathbf{H}^{(1)},\mathbf{H}^{(2)},\mathbf{H}^{(3)},\mathbf{H}^{(4)},
\mathbf{H}^{(-)}
\right]
\in \mathbb{R}^{5 \times d \times d_h},
\label{eq:generated_embedding_collection}
\end{equation}
where \(\mathbf{H}^{(1)}, \mathbf{H}^{(2)}, \mathbf{H}^{(3)},
\mathbf{H}^{(4)}\), and \(\mathbf{H}^{(-)}\) denote the embedding matrices
corresponding to the feature maps \(\mathbf{m}_1, \mathbf{m}_2,
\mathbf{m}_3, \mathbf{m}_4\), and \(\mathbf{m}^{-}\), respectively.

Combining the anchor encoder embedding matrix with the generated encoder embedding matrices
gives
\begin{equation}
\mathcal{H}
=
\left[
\mathbf{H}^{(a)},\mathbf{H}^{(1)},\mathbf{H}^{(2)},\mathbf{H}^{(3)},
\mathbf{H}^{(4)},\mathbf{H}^{(-)}
\right]
\in \mathbb{R}^{6 \times d \times d_h},
\label{eq:stacked_encoder_output}
\end{equation}
which forms the input to the projector. Thus, each embedding matrix in
\(\mathcal{H}\) corresponds to the anchor
feature map or one of the generated feature maps, and row \(j\) within that
embedding matrix gives the latent embedding of feature \(j\).

These encoder feature embeddings are next passed to the projector, which
maps them into the contrastive embedding space used by the training objective
in Section~\ref{subsubsec:opti}.

\subsubsection{Projector}
\label{subsubsec:projector}
This subsection describes the projector that maps the encoder output
\(\mathcal{H}\) in Eq.~\ref{eq:stacked_encoder_output} into the contrastive
embedding space. Within the proposed framework, the projector plays a dual
role: its normalized outputs define the space in which the InfoNCE objective
is optimized, while its unnormalized outputs retain the embedding magnitudes
that are later used for feature ranking. This separation is important because
the encoder output \(\mathcal{H}\) captures latent feature structure, whereas
the projector output defines the representation space used both for
contrastive optimization and for the final norm-based saliency criterion.

Formally, each embedding matrix in \(\mathcal{H}\) is passed through the
projector to produce a corresponding projector embedding matrix. When the
projector is applied to the anchor embedding matrix \(\mathbf{H}^{(a)}\), it
produces
\begin{equation}
\mathbf{Z}^{(a)} \in \mathbb{R}^{d \times d_z},
\label{eq:anchor_projector_matrix}
\end{equation}
where \(\mathbf{Z}^{(a)}\) denotes the anchor projector embedding matrix.
Likewise, when the projector is applied to the generated embedding collection
\(\bar{\mathbf{H}}\) in Eq.~\ref{eq:generated_embedding_collection}, it
produces the generated projector embedding collection
\begin{equation}
\bar{\mathbf{Z}}
=
\left[
\mathbf{Z}^{(1)},\mathbf{Z}^{(2)},\mathbf{Z}^{(3)},\mathbf{Z}^{(4)},
\mathbf{Z}^{(-)}
\right]
\in \mathbb{R}^{5 \times d \times d_z},
\label{eq:generated_projector_collection}
\end{equation}
where \(\mathbf{Z}^{(1)}, \mathbf{Z}^{(2)}, \mathbf{Z}^{(3)},
\mathbf{Z}^{(4)}\), and \(\mathbf{Z}^{(-)}\) denote the projector embedding
matrices corresponding to \(\mathbf{H}^{(1)}, \mathbf{H}^{(2)},
\mathbf{H}^{(3)}, \mathbf{H}^{(4)}\), and \(\mathbf{H}^{(-)}\),
respectively.

Combining the anchor projector embedding matrix with the generated projector
embedding matrices gives
\begin{equation}
\mathcal{Z}
=
\left[
\mathbf{Z}^{(a)},\mathbf{Z}^{(1)},\mathbf{Z}^{(2)},\mathbf{Z}^{(3)},
\mathbf{Z}^{(4)},\mathbf{Z}^{(-)}
\right]
\in \mathbb{R}^{6 \times d \times d_z},
\label{eq:stacked_projector_output}
\end{equation}
which forms the stacked projector embedding collection used by the contrastive
objective, where row \(j\) in each projector embedding matrix gives the
\(d_z\)-dimensional embedding of feature \(j\).

The projector is implemented as a residual multilayer perceptron composed of
two projector blocks followed by a final linear projection. Each projector
block consists of a linear transformation, batch normalization, and ReLU
activation. For an input embedding matrix \(\mathbf{H}^{(\mathbf{V})}\), the
first projector block maps each row into an intermediate projector hidden
representation of dimension \(d_p\), written as
\begin{equation}
\mathbf{U}^{(1)} \in \mathbb{R}^{d \times d_p},
\label{eq:projector_hidden_1}
\end{equation}
and the second projector block maps \(\mathbf{U}^{(1)}\) to a second hidden
representation
\begin{equation}
\mathbf{U}^{(2)} \in \mathbb{R}^{d \times d_p}.
\label{eq:projector_hidden_2}
\end{equation}
The residual projector representation is then formed as
\begin{equation}
\mathbf{U}^{\mathrm{res}} = \mathbf{U}^{(1)} + \mathbf{U}^{(2)},
\label{eq:projector_residual}
\end{equation}
and this summed representation is finally mapped by a linear layer to the
projector output space \(\mathbb{R}^{d \times d_z}\). Thus, the projector
first lifts the encoder embeddings into a hidden projector space of dimension
\(d_p\) before producing the final projector embedding matrix
\(\mathbf{Z}^{(\mathbf{V})} \in \mathbb{R}^{d \times d_z}\). The layer-wise
architecture is summarized in Table~\ref{tab:arch}.

\begin{table}[t]
\centering
\caption{Architecture of Component III (Neural Model). The encoder applies a
self-attention layer followed by two row-wise linear transformations, where the
first is followed by batch normalization, LeakyReLU activation, and dropout.
The projector is a residual multilayer perceptron in which each projector block
consists of a linear layer followed by batch normalization and ReLU
activation.}
\label{tab:arch}
\scriptsize
\setlength{\tabcolsep}{6pt}
\renewcommand{\arraystretch}{0.95}
\begin{tabular}{lll}
\toprule
Layer & Input & Output \\
\midrule
Input feature map & $\mathbf{V} \in \mathbb{R}^{d \times n}$ & $\mathbb{R}^{d \times n}$ \\
Self-attention & $\mathbb{R}^{d \times n}$ & $\mathbb{R}^{d \times n}$ \\
Encoder linear layer 1 & $\mathbb{R}^{d \times n}$ & $\mathbf{E}^{(\mathbf{V})} \in \mathbb{R}^{d \times d_e}$ \\
Encoder linear layer 2 & $\mathbf{E}^{(\mathbf{V})} \in \mathbb{R}^{d \times d_e}$ & $\mathbf{H}^{(\mathbf{V})} \in \mathbb{R}^{d \times d_h}$ \\
Projector block 1 & $\mathbf{H}^{(\mathbf{V})} \in \mathbb{R}^{d \times d_h}$ & $\mathbf{U}^{(1)} \in \mathbb{R}^{d \times d_p}$ \\
Projector block 2 & $\mathbf{U}^{(1)} \in \mathbb{R}^{d \times d_p}$ & $\mathbf{U}^{(2)} \in \mathbb{R}^{d \times d_p}$ \\
Residual add & $\mathbf{U}^{(1)}, \mathbf{U}^{(2)} \in \mathbb{R}^{d \times d_p}$ & $\mathbf{U}^{\mathrm{res}} \in \mathbb{R}^{d \times d_p}$ \\
Projector output layer & $\mathbf{U}^{\mathrm{res}} \in \mathbb{R}^{d \times d_p}$ & $\mathbf{Z}^{(\mathbf{V})} \in \mathbb{R}^{d \times d_z}$ \\
\bottomrule
\end{tabular}
\end{table}

The encoder-projector architecture described above is instantiated at two
different scales depending on the sample size of the dataset. This design
choice reflects the fact that, in the proposed feature-wise formulation, the
effective input dimensionality and the amount of available training signal are
both directly influenced by the number of samples\textbf{CHECK this probably should come before you check else we decide later where to keep}.

\subsubsection{Sample-Aware Model Scaling}
\label{subsubsec:capacity_regimes}
In the proposed model, the effective input dimensionality depends directly on
the number of samples available in a dataset, since each feature is
represented by its sample-profile vector. Prior work in contrastive self-supervised learning has shown that
representation quality is sensitive to architectural choices, and that
stronger performance can often be obtained with larger models and richer
optimization settings~\cite{chen2020simple,cao2021rethinking}. At
the same time, when the number of available training instances is limited,
excessive architectural capacity can make optimization less data-efficient
and increase the risk of over-parameterization.

Motivated by this tradeoff, we examined architectural choices across the
benchmark suite and observed a consistent sample-count-dependent trend in our
preliminary experiments: smaller-sample datasets favored lighter encoder and
projector configurations, whereas larger-sample datasets benefited from
larger ones. We therefore adopt two fixed sample-aware architectural regimes,
which help reduce underfitting for larger-sample datasets while avoiding
unnecessary over-parameterization for smaller-sample datasets. The exact
architectural configurations for the two regimes are provided in
Section~\ref{sec:implementation_details}.
\subsubsection{Optimization Objective}
\label{subsubsec:opti}
Having defined the encoder and projector feature embeddings, we
now specify the objective used to train the proposed framework. The training
objective combines a contrastive term,
\(\mathcal{L}_{\mathrm{con}}^{(m)}\), which encourages consistency between
the anchor map and each masked positive map while distinguishing them from
mismatched structure, with an inter-feature decorrelation term,
\(\mathcal{L}_{\mathrm{decorr}}\), which discourages different features from
collapsing to overly similar projector-space representations.

\paragraph{Contrastive Loss}
We optimize the proposed framework with an InfoNCE-based contrastive loss,
which is a standard objective for learning aligned representations from
multiple views of the same underlying instance while separating them from
mismatched instances~\cite{oord2018representation}. In the present feature-wise
setting, the anchor map and the masked positive maps provide multiple views of
the same feature, while the shuffled map provides an explicit mismatched view.
This choice is also supported by recent findings on self-supervised
representation learning. Prior work has shown that even when similarities are
computed on normalized embeddings, contrastive optimization still affects the
magnitudes and gradient behavior of the underlying raw
representations~\cite{wang2017normface,zhang2020spherical}. More recently,
Draganov et al.~\cite{draganovimportance} showed that InfoNCE-based training
systematically increases embedding norms and argued that larger norms are often
associated with more reinforced or more prototypical instances in the learned
representation space. Since our final feature ranking is derived from
projector-space embedding norm, we optimize the model with an InfoNCE-style
objective in order to utilize this norm-sensitive contrastive behavior as a
feature-saliency signal.

From Section~\ref{subsubsec:projector}, the projector produces the stacked
projector output \(\mathcal{Z}\) shown in
Eq.~\ref{eq:stacked_projector_output}, where
\(\mathbf{Z}^{(a)} \in \mathbb{R}^{d \times d_z}\) is the projector embedding
matrix corresponding to the anchor map \(\mathbf{X}^{\top}\),
\(\mathbf{Z}^{(1)},\mathbf{Z}^{(2)},\mathbf{Z}^{(3)},\mathbf{Z}^{(4)}\) are
the projector embedding matrices corresponding to the masked positive feature
maps \(\mathbf{m}_1,\mathbf{m}_2,\mathbf{m}_3,\mathbf{m}_4\), respectively,
and \(\mathbf{Z}^{(-)}\) is the projector embedding matrix corresponding to
the shuffled negative feature map \(\mathbf{m}^{-}\).

For each positive-view index \(m \in \{1,2,3,4\}\), let
\(\hat{\mathbf{Z}}^{(a)}\), \(\hat{\mathbf{Z}}^{(m)}\), and
\(\hat{\mathbf{Z}}^{(-)}\) denote the row-wise \(\ell_2\)-normalized versions
of \(\mathbf{Z}^{(a)}\), \(\mathbf{Z}^{(m)}\), and \(\mathbf{Z}^{(-)}\),
respectively. Then, for feature \(j\), the corresponding normalized
projector-space embeddings are
\(\hat{\mathbf{z}}^{(a)}_j\), \(\hat{\mathbf{z}}^{(m)}_j\), and
\(\hat{\mathbf{z}}^{(-)}_j\).

For each masked positive-view index \(m \in \{1,\dots,4\}\), we optimize an
InfoNCE-style objective~\cite{oord2018representation} given by
\begin{equation}
\mathcal{L}_{\mathrm{con}}^{(m)}
=
-\frac{1}{d}\sum_{j=1}^{d}
\log
\frac{
\exp\!\left(
\left\langle \hat{\mathbf{z}}^{(a)}_j,\hat{\mathbf{z}}^{(m)}_j \right\rangle / \tau
\right)
}{
\sum_{\ell=1}^{d}
\exp\!\left(
\left\langle \hat{\mathbf{z}}^{(a)}_j,\hat{\mathbf{z}}^{(m)}_\ell \right\rangle / \tau
\right)
+
\exp\!\left(
\left\langle \hat{\mathbf{z}}^{(a)}_j,\hat{\mathbf{z}}^{(-)}_j \right\rangle / \tau
\right)
}.
\label{eq:contrastive_loss}
\end{equation}
Here, \((\hat{\mathbf{z}}^{(a)}_j,\hat{\mathbf{z}}^{(m)}_j)\) forms the
positive pair for feature \(j\), the embeddings of all other features in the
current masked map act as implicit negatives through the summation over
\(\ell\), and the shuffled embedding \(\hat{\mathbf{z}}^{(-)}_j\) contributes
an additional explicit negative. This objective encourages each anchor feature
embedding to align with the corresponding embedding of its masked positive
view while separating it from mismatched structure, thereby favoring
perturbation-consistent projector-space representations.

\paragraph{Inter-Feature Decorrelation Regularizer.}
Contrastive consistency alone does not guarantee distinct projector-space
representations. In particular, correlated features may remain close in the
learned representation space and therefore yield similar saliency scores. To
discourage this, we introduce an inter-feature decorrelation regularizer on
the normalized anchor projector embeddings. Let
\begin{equation}
\hat{\mathbf{Z}}^{(a)} \in \mathbb{R}^{d \times d_z}
\label{eq:normalized_anchor_projector}
\end{equation}
denote the normalized anchor projector embedding matrix for all \(d\)
features, and let
\begin{equation}
\mathbf{C}^{(a)}=\hat{\mathbf{Z}}^{(a)}\hat{\mathbf{Z}}^{(a)\top}
\label{eq:anchor_similarity_matrix}
\end{equation}
denote the corresponding inter-feature cosine-similarity matrix. The
regularizer is defined as
\begin{equation}
\mathcal{L}_{\mathrm{decorr}}
=
\frac{1}{d^2}\left\|\mathbf{C}^{(a)}-\mathbf{I}_d\right\|_F^2,
\label{eq:decorrelation_loss}
\end{equation}
which suppresses off-diagonal similarities between distinct projector
embeddings while preserving unit self-similarity along the diagonal. In this
way, it encourages each feature to occupy a more distinct location in
projector space.

\paragraph{Total Loss}
In implementation, the decorrelation regularizer in
Eq.~\ref{eq:decorrelation_loss} is applied alongside each positive-view
contrastive term. Within each optimization step, the four positive views are
processed sequentially, and gradients from the corresponding four loss terms
are accumulated across the full positive-view loop. For each positive view,
the model recomputes the anchor projector embedding matrix
\(\mathbf{Z}^{(a)}\), the shuffled-negative projector embedding matrix
\(\mathbf{Z}^{(-)}\), and the current masked positive projector embedding
matrix \(\mathbf{Z}^{(m)}\). A single optimizer update is then performed
after the complete four-view loop. The resulting training objective
is
\begin{equation}
\mathcal{L}
=
\sum_{m=1}^{4}
\left(
\frac{1}{4}\mathcal{L}_{\mathrm{con}}^{(m)}
+
\lambda_{\mathrm{decorr}}\mathcal{L}_{\mathrm{decorr}}
\right),
\label{eq:total_loss}
\end{equation}
where \(\lambda_{\mathrm{decorr}}\) controls the contribution of the
decorrelation regularizer relative to the contrastive term. Thus, the total
objective in Eq.~\ref{eq:total_loss} introduces a dataset-dependent balance
between contrastive alignment and inter-feature separation through
\(\lambda_{\mathrm{decorr}}\). Since the strength of this regularization
affects how aggressively the model discourages similarity among feature
embeddings in \(\mathbf{Z}^{(a)}\), its setting must be matched to the scale
of the dataset and the available training signal.

\subsubsection{Sample-Aware Decorrelation Scaling}
\label{subsubsec:decorr_regimes}
The effect of the decorrelation penalty was also examined across datasets of
different sample sizes in our preliminary experiments. We observed that
smaller-sample datasets, such as ALLAML (\(n=72\)) and PROSTATE (\(n=102\)),
favored a weaker decorrelation weight, whereas larger-sample datasets, such as
COIL20 (\(n=1440\)) and BASEHOCK (\(n=1993\)), benefited from a stronger
one. Based on this empirical trend, we use two fixed sample-aware
decorrelation regimes. This choice helps avoid over-penalizing feature
similarity when the number of available training instances is limited, while
providing stronger redundancy control on datasets with substantially larger
sample sizes. The exact values used for the two regimes are given in
Section~\ref{sec:implementation_details}.


\subsection{Component IV: Selector}
This component describes how the learned projector-side representations are
converted into the final selected feature subset. It consists of two stages:
the \textit{Feature Selection Criterion}, which provides the primary global
ordering of features through projector-norm ranking, and
\textit{Laplacian-Gated Ranking Correction (LGRC)}, which performs a final
post hoc locality-preservation pass on the highest-ranked candidates. Thus,
the projector-norm ranking provides the main selection signal, while LGRC
acts only as a last-mile correction before the final subset is returned.

\subsubsection{Feature Selection Criterion}
\label{subsubsec:rank}
This subsection introduces the first-stage ranking criterion used to extract an
initial global ordering of features from the trained projector outputs. The
ranking is derived from the anchor-side projector output because the anchor
feature map \(\mathbf{X}^{\top}\) in Eq.~\ref{eq:inverted_map} corresponds to
the original transposed feature representation used by the proposed framework.
This choice is supported by prior work showing that, even when self-supervised
contrastive objectives are computed on normalized representations, the
magnitudes of the underlying embeddings still influence optimization dynamics
and can encode model confidence or prototypicality. In particular, Draganov
et al.~\cite{draganovimportance} showed that InfoNCE-based training
systematically grows embedding norms and that higher norms are often
associated with more reinforced or denser regions of the learned
representation space. Related work has likewise connected embedding magnitude
to confidence- and uncertainty-related behavior in representation
learning~\cite{scott2021von,kirchhof2023probabilistic}.

From the stacked projector output \(\mathcal{Z}\) in
Eq.~\ref{eq:stacked_projector_output}, we retain the anchor projector
embedding matrix \(\mathbf{Z}^{(a)} \in \mathbb{R}^{d \times d_z}\) introduced
in Eq.~\ref{eq:anchor_projector_matrix}. Its \(j\)-th row,
\(\mathbf{z}^{(a)}_j \in \mathbb{R}^{d_z}\), denotes the final projector
embedding of feature \(j\). We then use the \(\ell_2\)-norm of the raw
anchor-side projector output as the initial saliency signal. For feature
\(j\), this norm is defined as
\begin{equation}
\|\mathbf{z}^{(a)}_j\|_2
=
\left(
\sum_{\ell=1}^{d_z}
\left(z^{(a)}_{j,\ell}\right)^2
\right)^{1/2},
\label{eq:rank_l2norm}
\end{equation}
and the corresponding saliency score is written as
\begin{equation}
s_j = \|\mathbf{z}^{(a)}_j\|_2.
\label{eq:rank_score}
\end{equation}

Sorting features in descending order of \(s_j\) yields the initial global
ranking
\begin{equation}
\widehat{\mathcal{R}} = (\hat{r}_1,\hat{r}_2,\dots,\hat{r}_d),
\label{eq:initial_ranking}
\end{equation}
where
\begin{equation}
s_{\hat{r}_1} \ge s_{\hat{r}_2} \ge \dots \ge s_{\hat{r}_d}.
\label{eq:ranking_order}
\end{equation}
Thus, \(\hat{r}_1\) is the feature index with the largest projector norm,
\(\hat{r}_2\) is the feature index with the second-largest projector norm, and
so on. As defined in Eq.~\ref{eq:initial_ranking}, this norm-based ordering is
learned from a single training run and serves as the candidate ranking passed
to LGRC.

\subsubsection{Laplacian-Gated Ranking Correction}
\label{subsubsec:lap}
The projector-norm ranking \(\widehat{\mathcal{R}}\) in
Eq.~\ref{eq:initial_ranking} is learned once and provides the primary global
ordering of all \(d\) features. Because this ordering is based on individual
embedding magnitudes, it does not explicitly assess how well each retained
feature preserves the local neighborhood structure of the input data. LGRC
therefore applies a post hoc, cardinality-specific correction to the
highest-ranked candidates without retraining the neural model. Under the
evaluation protocol in Section~\ref{sec:eval}, LGRC is executed independently
for each target cardinality
\(k \in \{50,100,150,200,250,300\}\).

For a target cardinality \(k\), LGRC receives the complete projector-norm
ranking \(\widehat{\mathcal{R}}=(\hat{r}_1,\ldots,\hat{r}_d)\), ordered by
the descending saliency scores in Eq.~\ref{eq:ranking_order}. From the top of
this ranking, we form the candidate pool
\begin{equation}
\mathcal{P}_k = \{\hat{r}_1,\hat{r}_2,\dots,\hat{r}_p\},
\label{eq:lgrc_candidate_pool}
\end{equation}
where \(p=|\mathcal{P}_k|\) is the candidate-pool size associated with
\(k\) and is controlled by a fixed pool multiplier \(\alpha \geq 1\). In
implementation, the realized integer-valued pool size is
\begin{equation}
p = \min\!\bigl(d,\mathrm{round}(\alpha k)\bigr).
\label{eq:lgrc_pool_size}
\end{equation}
The remaining features
\(\{\hat{r}_{p+1},\hat{r}_{p+2},\dots,\hat{r}_d\}\) retain their original
order and form the reserve list from which replacements are drawn.

We compute a Laplacian score \(\ell_j\) for every feature \(j\) from the input
matrix \(\mathbf{X}\)~\cite{he2005laplacian}, where a lower score indicates
stronger preservation of the local sample-neighborhood structure. Let
\(\mathcal{L}_{\mathrm{fin}}=\{\ell_j:\ell_j \text{ is finite},\; j=1,\dots,d\}\)
denote the set of finite Laplacian scores. Given a fixed percentile level
\(q\), the gating threshold is defined by
\begin{equation}
\gamma_q = \operatorname{Quantile}_{q}\!\left(\mathcal{L}_{\mathrm{fin}}\right),
\label{eq:lgrc_threshold}
\end{equation}
with the convention that \(\gamma_q=+\infty\) if \(\mathcal{L}_{\mathrm{fin}}\)
is empty. At each iteration, the feature in \(\mathcal{P}_k\) with the largest
Laplacian score is identified. If this score exceeds \(\gamma_q\), the feature
is removed and replaced by the next available feature from the reserve list. The updated
pool is evaluated again, and this process continues until its largest retained
Laplacian score does not exceed \(\gamma_q\), or until the reserve list is
exhausted.

After the correction step, the features retained in \(\mathcal{P}_k\) are
restored to their original descending projector-norm order in
\(\widehat{\mathcal{R}}\). The top \(k\) entries of this corrected pool form
the final selected subset
\begin{equation}
\mathcal{S}_k = \{r^{(k)}_1,r^{(k)}_2,\dots,r^{(k)}_k\}.
\label{eq:lgrc_final_subset}
\end{equation}
Thus, LGRC retains projector norm as the primary saliency criterion while
preventing candidates with poor locality preservation from dominating the
selected subset. Algorithm~\ref{alg:lap_refine} summarizes the complete
procedure. The pool multiplier \(\alpha\), percentile level \(q\), and
internal Laplacian graph-construction settings are specified in
Section~\ref{sec:implementation_details}.

\begin{algorithm}[t]
\caption{Laplacian-Gated Ranking Correction}
\label{alg:lap_refine}
\KwIn{Input data matrix $\mathbf{X}$, initial projector-norm ranking $\widehat{\mathcal{R}}=(\hat{r}_1,\dots,\hat{r}_d)$, target cardinality $k$, pool multiplier $\alpha$, percentile level $q$}
\KwOut{Refined selected subset $\mathcal{S}_k$}

Set the candidate-pool size as $p = \min(d,\mathrm{round}(\alpha k))$\;

Construct the ranked candidate set $\mathcal{P}_k = \{\hat{r}_1,\hat{r}_2,\dots,\hat{r}_p\}$\;

Compute a Laplacian score $\ell_j$ for each original feature using $\mathbf{X}$\;

Collect the finite Laplacian scores into $\mathcal{L}_{\mathrm{fin}}$\;

Set $\gamma_q$ to the $q$-quantile of $\mathcal{L}_{\mathrm{fin}}$; if $\mathcal{L}_{\mathrm{fin}}$ is empty, set $\gamma_q = +\infty$\;

Set the reserve-pool pointer to the first ranked feature outside $\mathcal{P}_k$\;

\While{reserve features remain available}{
    Identify the feature in $\mathcal{P}_k$ with the largest Laplacian score\;
    
    \If{its score is no larger than $\gamma_q$}{
        \textbf{break}\;
    }
    
    Remove the worst-scoring feature from $\mathcal{P}_k$\;
    
    Insert the next available feature from $\widehat{\mathcal{R}}$ that is not already in $\mathcal{P}_k$\;
    
    Advance the reserve-pool pointer\;
}

Reorder the corrected pool $\mathcal{P}_k$ according to the original descending projector-norm order in $\widehat{\mathcal{R}}$\;

Return the top $k$ features of the reordered set as $\mathcal{S}_k$\;
\end{algorithm}

\section{Experiments}
\label{sec:experiments}
In this Section, we describe the evaluation protocol used to assess the feature
subsets selected by ICLFS and the baseline methods against which it is
compared followed by the implementation details and 
experimental settings.

\subsection{Evaluation Protocol}
\label{sec:eval}
Following prior UFS work~\cite{li2012unsupervised,shaham2022deep}, we evaluate
the feature selection quality of ICLFS through downstream clustering
performance. Specifically, we follow the evaluation protocol used in
LS-CAE~\cite{shaham2022deep}. For each baseline, we select feature subsets with
cardinality
\[
a \in \{50,100,150,200,250,300\},
\]
run $k$-means clustering~\cite{macqueen1967some} on the selected features 20
times, and compute clustering accuracy for each run. Clustering accuracy is
measured after optimally matching cluster assignments to ground-truth labels
using the Hungarian algorithm~\cite{munkres1957algorithms}. For each feature
cardinality $a$, we calculate the mean and standard deviation of clustering
accuracy across the 20 runs. Finally, for each dataset, we report the highest
mean clustering accuracy among the six evaluated feature subset sizes, with the
corresponding feature count shown in parentheses. All baseline methods are
evaluated under the same protocol.


\subsection{Datasets}
We evaluate ICLFS on twelve benchmark datasets drawn from the
\texttt{scikit-feature} repository.\footnote{\url{https://jundongl.github.io/scikit-feature/datasets.html}}
The benchmark spans image, biomedical, text, and mass-spectrometry domains and
follows a widely used UFS evaluation suite. Across these datasets, the number
of samples ranges from 60 to 7000, while the number of features ranges from
1024 to 10{,}000. The benchmark also includes both binary-class and
multi-class settings across heterogeneous domains, thereby providing a diverse
test bed for evaluating whether a feature selection method generalizes across
different structural regimes. A summary of the datasets is given in
Table~\ref{tab:datasets}.

\begin{table}[h]
  \caption{Summary of benchmark datasets.}
  \label{tab:datasets}
  \centering
  \begin{tabular}{lcccl}
    \toprule
    Dataset & Samples & Features & Classes & Domain \\
    \midrule
    ALLAML     & 72   & 7129  & 2  & Biological \\
    ARCENE     & 200  & 10000 & 2  & Mass spectrometry \\
    BASEHOCK   & 1993 & 4862  & 2  & Text \\
    COIL20     & 1440 & 1024  & 20 & Image \\
    GISETTE    & 7000 & 5000  & 2  & Handwritten digit / text-like \\
    LUNG       & 203  & 3312  & 5  & Biological \\
    NCI9       & 60   & 9712  & 9  & Biological \\
    PCMAC      & 1943 & 3289  & 2  & Text \\
    PROSTATE   & 102  & 5966  & 2  & Biological \\
    RELATHE    & 1427 & 4322  & 2  & Text \\
    TOX171     & 171  & 5748  & 4  & Biological / toxicology \\
    WARPPIE10P & 210  & 2420  & 10 & Image \\
    \bottomrule
  \end{tabular}
\end{table}

\subsection{Implementation Details}
\label{sec:implementation_details}
Our method is implemented in PyTorch and trained for 100 epochs using the Adam
optimizer with learning rate \(10^{-3}\) and weight decay \(10^{-4}\). The
self-attention module uses a single head, and the contrastive temperature is
fixed at \(\tau=0.05\). For masked-view generation, we use a light-mask keep
ratio of \(0.90\), a heavy-mask keep ratio of \(0.60\), and a complementary
masked pair constructed with keep ratio \(0.50\) and overlap ratio \(0.10\).

Following the sample-aware architectural and decorrelation regimes described
in Sections~\ref{subsubsec:capacity_regimes} and
\ref{subsubsec:decorr_regimes}, we use two sample-size-aware training
regimes:
\begin{itemize}
\item \textbf{Lower-sample regime} \(\mathcal{R}_{\mathrm{small}}\): this
regime is used for ALLAML, ARCENE, LUNG, NCI9, PROSTATE, TOX171, and
WARPPIE10P. It uses encoder hidden dimension \(d_e=16\), encoder embedding
dimension \(d_h=512\), projector hidden dimension \(d_p=128\), projector
output dimension \(d_z=16\), and decorrelation weight
\(\lambda_{\mathrm{decorr}}=0.20\).

\item \textbf{Higher-sample regime} \(\mathcal{R}_{\mathrm{large}}\): this
regime is used for BASEHOCK, COIL20, GISETTE, PCMAC, and RELATHE. It uses
encoder hidden dimension \(d_e=64\), encoder embedding dimension \(d_h=1440\),
projector hidden dimension \(d_p=2048\), projector output dimension
\(d_z=32\), and decorrelation weight
\(\lambda_{\mathrm{decorr}}=0.40\).
\end{itemize}

For the post hoc Laplacian-Gated Ranking Correction stage, we use a
candidate-pool multiplier of \(\alpha = 1.5\) and a percentile level of
\(q = 0.75\) for the gating threshold \(\gamma_q\). The affinity graph is
constructed with a \(3\)-nearest-neighbor cosine-distance graph and a
heat-kernel weighting scheme, where the kernel bandwidth is set from the mean
nonzero neighborhood distance. Each optimization step operates on the full
inverted data matrix \(\mathbf{X}^{\top} \in \mathbb{R}^{d \times n}\), so all
\(d\) feature instances are processed jointly in a single batch without
shuffling. To reduce memory usage, the four positive-view terms are processed
sequentially within each optimization step: for each view, the model
recomputes the anchor, shuffled negative, and current positive-view embeddings,
and gradients are accumulated across the full four-view loop before a single
optimizer update is applied. Experiments were run on an NVIDIA
RTX 5060 Ti with 16 GB of GPU memory, an AMD Ryzen 5 3600X CPU, and 8 GB of
system RAM. Code is available at.

\begin{table}[h]
  \caption{Dataset-specific architectural and decorrelation regimes used by ICLFS.}
  \label{tab:regimes}
  \centering
  \scriptsize
  \setlength{\tabcolsep}{5pt}
  \renewcommand{\arraystretch}{1.05}
  \begin{tabular}{lcc}
    \toprule
    Dataset & Regime & Settings \\
    \midrule
    ALLAML     & $\mathcal{R}_{\text{small}}$ & $d_e=16,\ d_h=512,\ d_p=128,\ d_z=16,\ \lambda_{\mathrm{decorr}}=0.20$ \\
    ARCENE     & $\mathcal{R}_{\text{small}}$ & $d_e=16,\ d_h=512,\ d_p=128,\ d_z=16,\ \lambda_{\mathrm{decorr}}=0.20$ \\
    LUNG       & $\mathcal{R}_{\text{small}}$ & $d_e=16,\ d_h=512,\ d_p=128,\ d_z=16,\ \lambda_{\mathrm{decorr}}=0.20$ \\
    NCI9       & $\mathcal{R}_{\text{small}}$ & $d_e=16,\ d_h=512,\ d_p=128,\ d_z=16,\ \lambda_{\mathrm{decorr}}=0.20$ \\
    PROSTATE   & $\mathcal{R}_{\text{small}}$ & $d_e=16,\ d_h=512,\ d_p=128,\ d_z=16,\ \lambda_{\mathrm{decorr}}=0.20$ \\
    TOX171     & $\mathcal{R}_{\text{small}}$ & $d_e=16,\ d_h=512,\ d_p=128,\ d_z=16,\ \lambda_{\mathrm{decorr}}=0.20$ \\
    WARPPIE10P & $\mathcal{R}_{\text{small}}$ & $d_e=16,\ d_h=512,\ d_p=128,\ d_z=16,\ \lambda_{\mathrm{decorr}}=0.20$ \\
    BASEHOCK   & $\mathcal{R}_{\text{large}}$ & $d_e=64,\ d_h=1440,\ d_p=2048,\ d_z=32,\ \lambda_{\mathrm{decorr}}=0.40$ \\
    COIL20     & $\mathcal{R}_{\text{large}}$ & $d_e=64,\ d_h=1440,\ d_p=2048,\ d_z=32,\ \lambda_{\mathrm{decorr}}=0.40$ \\
    GISETTE    & $\mathcal{R}_{\text{large}}$ & $d_e=64,\ d_h=1440,\ d_p=2048,\ d_z=32,\ \lambda_{\mathrm{decorr}}=0.40$ \\
    PCMAC      & $\mathcal{R}_{\text{large}}$ & $d_e=64,\ d_h=1440,\ d_p=2048,\ d_z=32,\ \lambda_{\mathrm{decorr}}=0.40$ \\
    RELATHE    & $\mathcal{R}_{\text{large}}$ & $d_e=64,\ d_h=1440,\ d_p=2048,\ d_z=32,\ \lambda_{\mathrm{decorr}}=0.40$ \\
    \bottomrule
  \end{tabular}
\end{table}
\subsection{Comparison with Existing Research Works}
We compare the performance of ICLFS against the following unsupervised feature
selection baselines: Laplacian Score (LS)~\cite{he2005laplacian},
Multi-Cluster Feature Selection (MCFS)~\cite{cai2010unsupervised},
Nonnegative Discriminative Feature Selection (NDFS)~\cite{li2012unsupervised},Spectral
Feature Selection (SPEC)~\cite{zhao2007spec},
(iv) Concrete Autoencoder (CAE)~\cite{balin2019concrete}, and
(v) LS-CAE~\cite{shaham2022deep}. To measure the performance of each method, we evaluated ICLFS and the baselines under the same protocol described in \ref{sec:eval}.

\section{Results and Analysis}
\label{sec:results}
In this Section, we compare ICLFS against existing UFS baselines and examine
the contribution of its main design components. The goal is to assess whether
the proposed feature-wise contrastive formulation yields more effective feature
rankings under the evaluation protocol described above in comparison to existing research works.

\subsection{Comparison with Existing Research Works}
Table~\ref{tab:main-results} reports the clustering accuracy (\%) achieved by
ICLFS and the competing methods across the evaluated feature subset sizes
\(a \in \{50,100,150,200,250,300\}\). Overall, ICLFS achieves the highest
clustering accuracy on 10 of the 12 benchmark datasets, a tied
second-highest accuracy on RELATHE, and a lower result on ALLAML. On
RELATHE, ICLFS attains 55.46\%, matching LS-CAE and remaining below the
best-performing baseline, CAE, at 55.87\%. On ALLAML, ICLFS attains
66.60\%, whereas the strongest result is obtained by LS-CAE at 71.11\%,
followed by MCFS and SPEC at 70.83\%.

Among the datasets on which ICLFS achieves the best result, the largest gains
over the strongest competing baseline are observed on PROSTATE, where CAE
attains 60.78\% and ICLFS reaches 83.97\%, on NCI9, where LS attains 41.08\%
and ICLFS reaches 52.33\%, and on LUNG, where MCFS attains 59.26\% and ICLFS
reaches 67.66\%. Additional improvements are observed on TOX171, where CAE
attains 50.20\% and ICLFS reaches 54.77\%; on WARPPIE10P, where SPEC attains
41.33\% and ICLFS reaches 44.14\%; on COIL20, where SPEC attains 66.83\% and
ICLFS reaches 68.68\%; on PCMAC, where CAE attains 51.31\% and ICLFS reaches
52.25\%; on GISETTE, where LS-CAE attains 75.98\% and ICLFS reaches 76.42\%;
on ARCENE, where NDFS and LS-CAE each attain 64.70\% and ICLFS reaches
65.17\%; and on BASEHOCK, where CAE attains 51.44\% and ICLFS reaches 51.88\%.

Dataset-wise, ICLFS attains the top result on ARCENE, BASEHOCK, COIL20,
GISETTE, LUNG, NCI9, PCMAC, PROSTATE, TOX171, and WARPPIE10P; the
tied second-highest result on RELATHE; and a lower result on ALLAML, where
LS-CAE, MCFS, and SPEC remain stronger. Across the full benchmark suite, the
strongest non-ICLFS baselines are LS-CAE on ALLAML and CAE on RELATHE. These
results indicate that the proposed inverted contrastive formulation provides a
strong and broadly effective alternative to existing unsupervised feature
selection methods across a diverse set of benchmark domains.

\begin{table*}[h]
\caption{Best clustering accuracy (\%) under the standard feature-count sweep. Each entry reports mean accuracy $\pm$ standard deviation, with the selected feature count in parentheses. Best is shown in bold and second best is underlined. NDFS on NCI9 is unavailable because it exceeded system RAM during baseline evaluation.}
  \label{tab:main-results}
  \centering
  \setlength{\tabcolsep}{4pt}
  \renewcommand{\arraystretch}{1.0}
  \resizebox{\textwidth}{!}{%
  \begin{tabular}{lccccccc}
    \toprule
    Dataset & LS & MCFS & NDFS & SPEC & CAE & LS-CAE & ICLFS \\
    \midrule
    ALLAML     & 63.61 $\pm$ 6.70 (150) & \underline{70.83 $\pm$ 2.74 (100)} & 67.08 $\pm$ 7.80 (200) & \underline{70.83 $\pm$ 4.26 (50)} & 54.51 $\pm$ 1.86 (200) & \textbf{71.11 $\pm$ 0.56 (300)} & 66.60 $\pm$ 1.20 (100) \\
    ARCENE     & 63.88 $\pm$ 2.74 (100) & 63.90 $\pm$ 2.14 (250) & \underline{64.70 $\pm$ 2.24 (100)} & 64.15 $\pm$ 2.23 (300) & 62.43 $\pm$ 1.73 (300) & \underline{64.70 $\pm$ 2.74 (300)} & \textbf{65.17 $\pm$ 1.21 (250)} \\
    BASEHOCK   & 50.87 $\pm$ 1.15 (300) & 51.21 $\pm$ 1.27 (250) & 50.68 $\pm$ 0.72 (100) & 50.94 $\pm$ 1.40 (100) & \underline{51.44 $\pm$ 1.12 (250)} & 50.95 $\pm$ 1.00 (50) & \textbf{51.88 $\pm$ 1.13 (100)} \\
    COIL20     & 56.31 $\pm$ 2.21 (300) & 60.78 $\pm$ 3.03 (200) & 57.86 $\pm$ 4.48 (300) & \underline{66.83 $\pm$ 2.85 (300)} & 64.07 $\pm$ 3.08 (250) & 64.76 $\pm$ 4.14 (250) & \textbf{68.68 $\pm$ 2.69 (150)} \\
    GISETTE    & 59.08 $\pm$ 6.24 (250) & 63.87 $\pm$ 9.14 (150) & 62.13 $\pm$ 6.20 (300) & 68.35 $\pm$ 0.10 (250) & 56.51 $\pm$ 0.04 (50) & \underline{75.98 $\pm$ 0.22 (200)} & \textbf{76.42 $\pm$ 0.04 (250)} \\
    LUNG       & 57.07 $\pm$ 4.20 (250) & \underline{59.26 $\pm$ 4.19 (100)} & 54.53 $\pm$ 5.65 (150) & 57.27 $\pm$ 5.59 (50) & 53.42 $\pm$ 5.21 (250) & 57.27 $\pm$ 4.77 (200) & \textbf{67.66 $\pm$ 7.17 (300)} \\
    NCI9       & \underline{41.08 $\pm$ 3.62 (300)} & 37.83 $\pm$ 4.09 (300) & -- & 40.33 $\pm$ 2.51 (250) & 39.08 $\pm$ 3.78 (300) & 37.17 $\pm$ 4.05 (250) & \textbf{52.33 $\pm$ 3.67 (100)} \\
    PCMAC      & 50.61 $\pm$ 0.28 (300) & 50.67 $\pm$ 0.47 (150) & 50.81 $\pm$ 0.70 (300) & 50.86 $\pm$ 0.54 (300) & \underline{51.31 $\pm$ 0.99 (100)} & 50.78 $\pm$ 0.74 (200) & \textbf{52.25 $\pm$ 1.33 (50)} \\
    PROSTATE   & 57.21 $\pm$ 0.47 (100) & 57.84 $\pm$ 0.00 (50) & 56.08 $\pm$ 0.39 (50) & 57.55 $\pm$ 0.45 (100) & \underline{60.78 $\pm$ 0.00 (50)} & 59.85 $\pm$ 0.58 (100) & \textbf{83.97 $\pm$ 0.47 (50)} \\
    RELATHE    & 54.23 $\pm$ 0.94 (100) & 54.94 $\pm$ 1.26 (250) & 54.57 $\pm$ 1.48 (300) & 54.38 $\pm$ 0.84 (150) & \textbf{55.87 $\pm$ 1.03 (250)} & 55.46 $\pm$ 1.06 (300) & \underline{55.46 $\pm$ 1.62 (200)} \\
    TOX171     & 49.09 $\pm$ 2.96 (150) & 47.31 $\pm$ 3.18 (200) & 47.08 $\pm$ 3.42 (200) & 48.45 $\pm$ 0.03 (50) & \underline{50.20 $\pm$ 3.77 (300)} & 49.77 $\pm$ 2.88 (50) & \textbf{54.77 $\pm$ 4.53 (100)} \\
    WARPPIE10P & 37.43 $\pm$ 3.08 (100) & 31.57 $\pm$ 2.74 (200) & 39.64 $\pm$ 3.26 (50) & \underline{41.33 $\pm$ 3.39 (50)} & 28.36 $\pm$ 1.47 (250) & 32.57 $\pm$ 2.02 (100) & \textbf{44.14 $\pm$ 3.02 (50)} \\
    \bottomrule
  \end{tabular}%
  }
\end{table*}

\subsection{Ablation Study}
We evaluate the contribution of the components of ICLFS through a
progressive ablation. Starting from the complete ICLFS model, we first remove
Laplacian-Gated Ranking Correction (LGRC) to obtain ICLFS-WL. We then
additionally remove the decorrelation regularizer, yielding ICLFS-WLD,
followed by disabling the attention layer to form ICLFS-WLDA. Finally, we
replace the structured four-view masking scheme with a single random-masked
positive view, resulting in ICLFS-WLDAR.
This progression defines a hierarchy of increasingly simplified variants and
allows us to isolate the effect of LGRC, decorrelation-aware
regularization, relational feature interaction, and structured multi-view
augmentation. Table~\ref{tab:ablation} reports a progressive ablation of the proposed
framework. Removing LGRC first (ICLFS-WL) causes only small
drops relative to the full model, namely 0.51\% on COIL20, 0.34\% on PCMAC,
0.64\% on PROSTATE, and 1.14\% on WARPPIE10P, while ARCENE remains unchanged.
This pattern indicates that the core contrastive ranking framework is already
robust, and that LGRC acts primarily as a final locality-aware
refiner rather than as the main source of performance.

In contrast, once ablations enter the core representation-learning pipeline,
the degradation becomes substantially larger. Removing the decorrelation
regularizer in addition to LGRC (ICLFS-WLD) reduces accuracy by 1.27\% on
ARCENE, 3.27\% on COIL20, 0.51\% on PCMAC, 14.51\% on PROSTATE, and 8.54\% on
WARPPIE10P relative to the full model. Disabling the attention layer as well
(ICLFS-WLDA) produces even larger drops on ARCENE (8.87\%), COIL20 (9.90\%),
PCMAC (1.12\%), and PROSTATE (12.21\%), while WARPPIE10P remains only slightly
below the full model by 0.83\%. Finally, replacing the structured four-view
masking design with a single random-masked positive view (ICLFS-WLDAR) causes
the largest declines on several datasets, including 6.17\% on ARCENE, 6.56\%
on COIL20, 26.91\% on PROSTATE, and 6.71\% on WARPPIE10P, although it yields a
local improvement of 1.01\% on PCMAC. Overall, these results show that the
largest performance losses arise when core components of the proposed
representation-learning pipeline are removed, thereby supporting the design
choice of structured multi-view contrastive learning, attention-based feature
interaction, and projector-space decorrelation as the main drivers of ICLFS.
\begin{table*}[h]
  \caption{Ablation study using best clustering accuracy (\%). `ICLFS-WL` removes LGRC, `ICLFS-WLD` removes the decorrelation regularizer in addition, `ICLFS-WLDA` further disables the attention layer, and `ICLFS-WLDAR` replaces the structured four-view masking scheme with a single random-masked positive view.}
  \label{tab:ablation}
  \centering
  \begin{tabular}{lccccc}
    \toprule
    Dataset & ICLFS & ICLFS-WL & ICLFS-WLD & ICLFS-WLDA & ICLFS-WLDAR\\
    \midrule
    ARCENE     & \textbf{65.17} & 65.17 & 63.90 & 56.30 & 59.00 \\
    COIL20     & \textbf{68.68} & 68.17 & 65.41 & 58.78 & 62.12 \\
    PCMAC      & 52.25 & 51.91 & 51.74 & 51.13 & \textbf{53.26} \\
    PROSTATE   & \textbf{83.97} & 83.33 & 69.46 & 71.76 & 57.06 \\
    WARPPIE10P & \textbf{44.14} & 43.00 & 35.60 & 43.31 & 37.43 \\
    \bottomrule
  \end{tabular}
\end{table*}

\subsection{Alternate Selection Criterion}
Our default feature ranking criterion uses the projector-side embedding norm
\[
s_j = \|\mathbf{z}^{(a)}_j\|_2.
\]
To assess whether the learned representations contain useful ranking signal
beyond this default choice, we compare it against several alternative scoring
rules derived from the encoder and projector outputs. In addition to
\emph{$z$-norm}, we evaluate \emph{$h$-norm}, which ranks features by the
encoder-space magnitude $\|\mathbf{h}^{(a)}_j\|_2$, as well as two fusion rules
that combine encoder-side and projector-side information.

Let
\[
\tilde{h}_j \in [0,1], \qquad \tilde{z}_j \in [0,1]
\]
denote the min-max normalized versions of $\|\mathbf{h}^{(a)}_j\|_2$ and
$\|\mathbf{z}^{(a)}_j\|_2$, respectively. We then define a weighted-sum criterion
\[
s_j^{\mathrm{ws}} = 0.35\,\tilde{h}_j + 0.65\,\tilde{z}_j,
\]
which places greater weight on the projector-side signal, a harmonic-fusion
criterion
\[
s_j^{\mathrm{harm}} =
\frac{2\,\tilde{h}_j \tilde{z}_j}{\tilde{h}_j + \tilde{z}_j + \epsilon},
\]
where $\epsilon$ is a small constant for numerical stability.

\begin{table*}[t]
  \caption{Best clustering accuracy (\%) under alternative feature scoring criteria on a representative 5-dataset subset. Best is in bold and second best is underlined. Tied best entries are all bolded.}
  \label{tab:alt-score}
  \centering
  \begin{tabular}{lcccc}
    \toprule
    Dataset & $z$-norm & $h$-norm & Weighted Sum & Harmonic-$hz$ \\
    \midrule
    ARCENE     & \textbf{65.17 $\pm$ 1.21 (250)} & 63.30 $\pm$ 0.71 (300) & \underline{64.35 $\pm$ 1.49 (100)} & 64.30 $\pm$ 1.45 (150) \\
    PCMAC      & \textbf{52.25 $\pm$ 1.33 (50)} & 50.91 $\pm$ 0.78 (250) & \underline{51.27 $\pm$ 0.57 (50)} & 51.06 $\pm$ 0.67 (100) \\
    WARPPIE10P & \textbf{44.14 $\pm$ 3.02 (50)} & 36.95 $\pm$ 3.01 (50) & \underline{43.36 $\pm$ 4.18 (50)} & 41.38 $\pm$ 2.61 (50) \\
    COIL20     & \textbf{68.68 $\pm$ 2.69 (150)} & 66.03 $\pm$ 2.82 (150) & \underline{66.91 $\pm$ 3.06 (150)} & 66.77 $\pm$ 2.71 (200) \\
    PROSTATE   & \underline{83.97 $\pm$ 0.47 (50)} & 81.57 $\pm$ 10.09 (50) & \textbf{84.31 $\pm$ 0.00 (50)} & 83.48 $\pm$ 1.43 (50) \\
    \bottomrule
  \end{tabular}
\end{table*}

Table~\ref{tab:alt-score} compares these alternative scoring rules across the
five evaluated datasets. The results show that the default projector-norm
criterion remains the strongest overall choice. In particular, $z$-norm
achieves the best clustering accuracy on ARCENE, PCMAC, WARPPIE10P, and
COIL20, indicating that the anchor-side projector embedding magnitude
provides the most stable ranking signal across these datasets. The margins are
especially clear on WARPPIE10P, where $z$-norm reaches 44.14\% compared with
43.36\% for Weighted Sum and 41.38\% for Harmonic-$hz$, and on COIL20, where
it reaches 68.68\% compared with 66.91\% and 66.77\%, respectively.

The main exception is PROSTATE, where the Weighted Sum rule achieves the best
result at 84.31\%, slightly exceeding the default $z$-norm score of 83.97\%.
This suggests that, on some datasets, encoder-side magnitude and
projector-side magnitude may provide complementary ranking information when
combined at the score level. However, this advantage is not consistent across
the remaining datasets, where the fused criteria either remain below the
default rule or only approach it marginally.

The $h$-norm criterion is the weakest overall, which indicates that encoder-side
magnitude alone does not provide as reliable a feature ranking signal as the
projector-side embedding norm learned under the contrastive objective. Overall,
these results support the robustness of the default projector-norm ranking
rule, while also showing that simple score-level fusion can occasionally offer
a modest benefit on specific datasets.

\section{Conclusion}
\label{sec:conclusion}
In this paper, we proposed Inverted Contrastive Learning for Unsupervised
Feature Selection (ICLFS), a feature-wise contrastive framework that
reformulates unsupervised feature selection as a representation learning
problem over features rather than samples. By inverting the data matrix,
constructing multiple masked positive views together with a shuffled negative
view, and optimizing the resulting feature-wise representations under an
InfoNCE-based objective, the proposed framework learns projector-space
embeddings whose magnitudes provide an effective saliency signal for feature
ranking. To further improve feature separation and reduce redundancy, we
combined this ranking mechanism with an inter-feature decorrelation
regularizer and a final Laplacian-Gated Ranking Correction stage.

Extensive experiments on 12 benchmark datasets showed that ICLFS achieves the
best clustering accuracy on 10 datasets against both classical and neural
baselines under the standard clustering-based UFS evaluation protocol, while
remaining competitive on the remaining two. These results indicate that
feature-wise contrastive representation consistency provides a strong and
broadly effective alternative to neighborhood-, cluster-, and
reconstruction-based formulations of unsupervised feature selection.

Future work will explore how this feature-wise representation learning
framework can be extended beyond the present ranking-based setting to support
other forms of feature selection, including alternative subset selection
criteria, task-adaptive selection strategies, and hybrid selection mechanisms
built directly on learned feature representations. More broadly, this
direction may help establish a more general representation-learning foundation
for unsupervised feature selection.

\section*{Declaration of Competing Interest}

\par The authors have no relevant financial or non-financial interests to disclose.

\bibliography{sn-bibliography}

@article{shaham2022deep,
  title={Deep unsupervised feature selection by discarding nuisance and correlated features},
  author={Shaham, Uri and Lindenbaum, Ofir and Svirsky, Jonathan and Kluger, Yuval},
  journal={Neural Networks},
  volume={152},
  pages={34--43},
  year={2022},
  publisher={Elsevier}
}

@article{he2005laplacian,
  title={Laplacian score for feature selection},
  author={He, Xiaofei and Cai, Deng and Niyogi, Partha},
  journal={Advances in neural information processing systems},
  volume={18},
  year={2005}
}

@inproceedings{cai2010unsupervised,
  title={Unsupervised feature selection for multi-cluster data},
  author={Cai, Deng and Zhang, Chiyuan and He, Xiaofei},
  booktitle={Proceedings of the 16th ACM SIGKDD international conference on Knowledge discovery and data mining},
  pages={333--342},
  year={2010}
}

@inproceedings{li2012unsupervised,
  title={Unsupervised feature selection using nonnegative spectral analysis},
  author={Li, Zechao and Yang, Yi and Liu, Jing and Zhou, Xiaofang and Lu, Hanqing},
  booktitle={Proceedings of the AAAI conference on artificial intelligence},
  volume={26},
  number={1},
  pages={1026--1032},
  year={2012}
}

@article{cai2018feature,
  title   = {Feature Selection in Machine Learning: A New Perspective},
  author  = {Cai, Jie and Luo, Jiawei and Wang, Shulin and Yang, Sheng},
  journal = {Neurocomputing},
  volume  = {300},
  pages   = {70--79},
  year    = {2018},
  issn    = {0925-2312},
  doi     = {10.1016/j.neucom.2017.11.077},
  url     = {https://www.sciencedirect.com/science/article/pii/S0925231218302911}
}

@article{saeys2007review,
  title   = {A Review of Feature Selection Techniques in Bioinformatics},
  author  = {Saeys, Yvan and Inza, I{\~n}aki and Larra{\~n}aga, Pedro},
  journal = {Bioinformatics},
  volume  = {23},
  number  = {19},
  pages   = {2507--2517},
  year    = {2007},
  doi     = {10.1093/bioinformatics/btm344},
  url     = {https://academic.oup.com/bioinformatics/article/23/19/2507/185254}
}

@inproceedings{draganovimportance,
  title={On the Importance of Embedding Norms in Self-Supervised Learning},
  author={Draganov, Andrew and Vadgama, Sharvaree and Damrich, Sebastian and B{\"o}hm, Jan Niklas and Maes, Lucas and Kobak, Dmitry and Bekkers, Erik J},
  booktitle={Forty-second International Conference on Machine Learning}
}

@inproceedings{wang2017normface,
  title={Normface: L2 hypersphere embedding for face verification},
  author={Wang, Feng and Xiang, Xiang and Cheng, Jian and Yuille, Alan Loddon},
  booktitle={Proceedings of the 25th ACM international conference on Multimedia},
  pages={1041--1049},
  year={2017}
}

@article{zhang2020spherical,
  title={Deep metric learning with spherical embedding},
  author={Zhang, Dingyi and Li, Yingming and Zhang, Zhongfei},
  journal={Advances in Neural Information Processing Systems},
  volume={33},
  pages={18772--18783},
  year={2020}
}

@inproceedings{kirchhof2023probabilistic,
  title={Probabilistic contrastive learning recovers the correct aleatoric uncertainty of ambiguous inputs},
  author={Kirchhof, Michael and Kasneci, Enkelejda and Oh, Seong Joon},
  booktitle={International Conference on Machine Learning},
  pages={17085--17104},
  year={2023},
  organization={PMLR}
}

@inproceedings{macqueen1967some,
  title={Multivariate observations},
  author={MacQueen, J},
  booktitle={Proceedings ofthe 5th Berkeley symposium on mathematical statisticsand probability},
  volume={1},
  pages={281--297},
  year={1967},
  organization={University of California press Oakland, CA, USA}
}

@article{munkres1957algorithms,
  title={Algorithms for the assignment and transportation problems},
  author={Munkres, James},
  journal={Journal of the society for industrial and applied mathematics},
  volume={5},
  number={1},
  pages={32--38},
  year={1957},
  publisher={SIAM}
}

@inproceedings{zhao2007spec,
  title={Spectral feature selection for supervised and unsupervised learning},
  author={Zhao, Zheng and Liu, Huan},
  booktitle={Proceedings of the 24th international conference on Machine learning},
  pages={1151--1157},
  year={2007}
}

@inproceedings{balin2019concrete,
  title     = {Concrete Autoencoders: Differentiable Feature Selection and Reconstruction},
  author    = {Bal{\i}n, Muhammed Fatih and Abid, Abubakar and Zou, James},
  booktitle = {Proceedings of the 36th International Conference on Machine Learning},
  pages     = {444--453},
  year      = {2019},
  volume    = {97},
  series    = {Proceedings of Machine Learning Research},
  publisher = {PMLR},
  url       = {https://proceedings.mlr.press/v97/balin19a.html}
}

@inproceedings{scott2021von,
  title={von mises-fisher loss: An exploration of embedding geometries for supervised learning},
  author={Scott, Tyler R and Gallagher, Andrew C and Mozer, Michael C},
  booktitle={Proceedings of the IEEE/CVF international conference on computer vision},
  pages={10612--10622},
  year={2021}
}

@article{li2017feature,
  title={Feature selection: A data perspective},
  author={Li, Jundong and Cheng, Kewei and Wang, Suhang and Morstatter, Fred and Trevino, Robert P and Tang, Jiliang and Liu, Huan},
  journal={ACM computing surveys (CSUR)},
  volume={50},
  number={6},
  pages={1--45},
  year={2017},
  publisher={ACM New York, NY, USA}
}

@article{guyon2003introduction,
  title={An introduction to variable and feature selection},
  author={Guyon, Isabelle and Elisseeff, Andr{\'e}},
  journal={Journal of machine learning research},
  volume={3},
  number={Mar},
  pages={1157--1182},
  year={2003}
}

@article{oord2018representation,
  title={Representation learning with contrastive predictive coding},
  author={Oord, Aaron van den and Li, Yazhe and Vinyals, Oriol},
  journal={arXiv preprint arXiv:1807.03748},
  year={2018}
}

@article{lindenbaum2021differentiable,
  title={Differentiable unsupervised feature selection based on a gated laplacian},
  author={Lindenbaum, Ofir and Shaham, Uri and Peterfreund, Erez and Svirsky, Jonathan and Casey, Nicolas and Kluger, Yuval},
  journal={Advances in neural information processing systems},
  volume={34},
  pages={1530--1542},
  year={2021}
}

@article{segalspectral,
  title={Spectral Self-supervised Feature Selection},
  author={Segal, Daniel and Lindenbaum, Ofir and Jaffe, Ariel},
  journal={Transactions on Machine Learning Research}
}

@inproceedings{chen2020simple,
  title={A simple framework for contrastive learning of visual representations},
  author={Chen, Ting and Kornblith, Simon and Norouzi, Mohammad and Hinton, Geoffrey},
  booktitle={International conference on machine learning},
  pages={1597--1607},
  year={2020},
  organization={PmLR}
}

@article{cao2021rethinking,
  title={Rethinking self-supervised learning: Small is beautiful},
  author={Cao, Yun-Hao and Wu, Jianxin},
  journal={arXiv preprint arXiv:2103.13559},
  year={2021}
}

@article{garcia2016high,
  title={High-dimensional feature selection via feature grouping: A Variable Neighborhood Search approach},
  author={Garc{\'\i}a-Torres, Miguel and G{\'o}mez-Vela, Francisco and Meli{\'a}n-Batista, Bel{\'e}n and Moreno-Vega, J Marcos},
  journal={Information Sciences},
  volume={326},
  pages={102--118},
  year={2016},
  publisher={Elsevier}
}

@article{urbanowicz2018relief,
  title={Relief-based feature selection: Introduction and review},
  author={Urbanowicz, Ryan J and Meeker, Melissa and La Cava, William and Olson, Randal S and Moore, Jason H},
  journal={Journal of biomedical informatics},
  volume={85},
  pages={189--203},
  year={2018},
  publisher={Elsevier}
}

\end{document}